\documentclass[10pt,twocolumn,letterpaper]{article}

\usepackage{style}
\usepackage{times}
\usepackage{epsfig}
\usepackage{graphicx}
\usepackage{amsmath}
\usepackage{amssymb}
\usepackage{gensymb}
\usepackage{authblk}
\usepackage{subcaption}

\usepackage[pagebackref=true,breaklinks=true,letterpaper=true,colorlinks,bookmarks=false]{hyperref}

\newif \ifdrawfigures
\drawfigurestrue
\newif \ifusevspace
\usevspacetrue
\newcommand{\mydraft}{false}

\newcommand{\twofigure}[4]{
{\hbox{\includegraphics[width=#1,draft=\mydraft]{#3}\hspace{#2}\includegraphics[width=#1,draft=\mydraft]{#4}}}
}

\cvprfinalcopy

\begin{document}

%%%%%%%%% TITLE
\title{Three Dimensional Reconstruction of Botanical Trees\\with Simulatable Geometry}

\setlength{\affilsep}{0.3cm}

\author[1,2]{Ed Quigley}
\author[1]{Winnie Lin}
\author[1]{Yilin Zhu}
\author[1,2]{Ronald Fedkiw}

\affil[1]{Stanford University}
\affil[2]{Industrial Light \& Magic}

\makeatletter
\renewcommand\AB@emaillist{\small{\texttt{\{equigley,winnielin,yilinzhu,rfedkiw\}@stanford.edu}}}
\makeatother

\maketitle

%%%%%%%%%%%%%%%%%%%%%%%%%%%%%%%%%%%%%%%%%%%%%%%%%%%%%%%%%%%%%%%%%%%%%%%%%%%%%%%
% Abstract
%%%%%%%%%%%%%%%%%%%%%%%%%%%%%%%%%%%%%%%%%%%%%%%%%%%%%%%%%%%%%%%%%%%%%%%%%%%%%%%
\begin{abstract}
We tackle the challenging problem of creating full and accurate three dimensional reconstructions of botanical trees with the topological and geometric accuracy required for subsequent physical simulation, e.g.\ in response to wind forces.
Although certain aspects of our approach would benefit from various improvements, our results exceed the state of the art especially in geometric and topological complexity and accuracy.
Starting with two dimensional RGB image data acquired from cameras attached to drones, we create point clouds, textured triangle meshes, and a simulatable and skinned cylindrical articulated rigid body model.
We discuss the pros and cons of each step of our pipeline, and in order to stimulate future research we make the raw and processed data from every step of the pipeline as well as the final geometric reconstructions publicly available.
\end{abstract}

%%%%%%%%%%%%%%%%%%%%%%%%%%%%%%%%%%%%%%%%%%%%%%%%%%%%%%%%%%%%%%%%%%%%%%%%%%%%%%%
% Introduction
%%%%%%%%%%%%%%%%%%%%%%%%%%%%%%%%%%%%%%%%%%%%%%%%%%%%%%%%%%%%%%%%%%%%%%%%%%%%%%%
\section{Introduction}
Human-inhabited outdoor environments typically contain ground surfaces such as grass and roads, transportation vehicles such as cars and bikes, buildings and structures, and humans themselves, but are also typically intentionally populated by a large number of trees and shrubbery; most of the motion in such environments comes from humans, their vehicles, and wind-driven plants/trees.
Tree reconstruction and simulation are obviously useful for AR/VR, architectural design and modeling, film special effects, etc.
For example, when filming actors running through trees, one would like to create virtual versions of those trees with which a chasing dinosaur could interact.
Other uses include studying roots and plants for agriculture~\cite{zheng:2011:detailed,estrada:2015:tree,fuentes:2017:robust} or assessing the health of trees especially in remote locations (similar in spirit to \cite{zuffi:2018:lions}).
2.5D data, i.e.\ 2D images with some depth information, is typically sufficient for robotic navigation, etc.; however, there are many problems that require true 3D scene understanding to the extent one could 3D print objects and have accurate geodesics.
Whereas navigating around objects might readily generalize into categories or strategies such as `move left,' `move right,' `step up,' `go under,' etc., the 3D object understanding required for picking up a cup, knocking down a building, moving a stack of bricks or a pile of dirt, or simulating a tree moving in the wind requires significantly higher fidelity.
As opposed to random trial and error, humans often use mental simulations to better complete a task, e.g.\ consider stacking a card tower, avoiding a falling object, or hitting a baseball (visualization is quite important in sports);
thus, physical simulation can play an important role in end-to-end tasks, e.g. see~\cite{kloss:2017:combining,peng:2017:deeploco,jiang:2018:data} for examples of combining simulation and learning.

Accurate 3D shape reconstruction is still quite challenging.
Recently, Malik argued\footnote{Jitendra Malik, Stanford cs231n guest lecture, 29 May 2018} that one should not apply general purpose reconstruction algorithms to say a car and a tree and expect both reconstructions to be of high quality.
Rather, he said that one should use domain-specific knowledge as he has done for example in~\cite{kanazawa:2018:learning}.
Another example of this specialization strategy is to rely on the prior that many indoor surfaces are planar in order to reconstruct office spaces~\cite{huang:2017:3dlite} or entire buildings~\cite{armeni:2016:3d,armeni:2017:joint}.
Along the same lines, \cite{zuffi:2018:lions} uses a base animal shape as a prior for their reconstructions of wild animals.
Thus, we similarly take a specialized approach using a generalized cylinder prior for both large and medium scale features.

In Section~\ref{sec:data}, we discuss our constraints on data collection as well as the logistics behind the choices we made for the hardware (cameras and drones) and software (structure from motion, multi-view stereo, inverse rendering, etc.) used to obtain our raw and processed data.
Sections~\ref{sec:sim}, \ref{sec:med_branches}, and \ref{sec:unresolved} then describe how we create geometry from that data with enough efficacy for physical simulation.
Section~\ref{sec:learning} discusses our use of machine learning, and Section~\ref{sec:exp} presents a number of experimental results.

%%%%%%%%%%%%%%%%%%%%%%%%%%%%%%%%%%%%%%%%%%%%%%%%%%%%%%%%%%%%%%%%%%%%%%%%%%%%%%%
% Previous Work
%%%%%%%%%%%%%%%%%%%%%%%%%%%%%%%%%%%%%%%%%%%%%%%%%%%%%%%%%%%%%%%%%%%%%%%%%%%%%%%
\section{Previous Work}

\textbf{Tree Modeling and Reconstruction:}
Researchers in computer graphics have been interested in modeling trees and plants for decades~\cite{lindenmayer:1968:mathematical,bloomenthal:1985:modeling,weber:1995:tree-model,prusinkiewicz:1997:visual,stava:2014:inverse}.
SpeedTree\footnote{\texttt{https://speedtree.com}} is probably the most popular software utilized, and their group has begun to consider the incorporation of data-driven methods. 
Amongst the data-driven approaches, \cite{tan:2007:image} is most similar to ours combining point cloud and image segmentation data to build coarse-scale details of a tree; however, they generate fine-scale details procedurally using a self-similarity assumption and image-space growth constraints, whereas we aim to capture more accurate finer structures from the image data.
Other data-driven approaches include \cite{livny:2010:tree-point-cloud} which automatically estimates skeletal structure of trees from point cloud data, \cite{xie:2015:tree} which builds tree models by assembling pieces from a database of scanned tree parts, etc.

Many of these specialized, data-driven approaches for trees are built upon more general techniques such as the traditional combination of structure from motion (see e.g.\ \cite{wu:2013:towards}) and multi-view stereo (see e.g.\ \cite{furukawa:2010:accurate}).
In the past, researchers studying 3D reconstruction have engineered general approaches to reconstruct fine details of small objects captured by sensors in highly controlled environments~\cite{seitz:2006:comparison}.
At the other end of the spectrum, researchers have developed approaches for reconstructing building- or even city-scale objects using large amounts of image data available online~\cite{agarwal:2009:building}.
Our goal is to obtain a 3D model of a tree with elements from both of these approaches: the scale of a large structure with the fine details of its many branches and twigs.
However, unlike in general reconstruction approaches, we cannot simply collect images online or capture data using a high-end camera.

To address similar challenges in specialized cases, researchers take advantage of domain-specific prior knowledge.
\cite{zhou:2008:circular} uses a generalized cylinder prior (similar to us) for reconstructing tubular structures observed during medical procedures and illustrates that this approach performs better than simple structure from motion.
The process of creating a mesh that faithfully reflects topology and subsequently refining its geometry is similar in spirit to \cite{xu:2018:monoperfcap}, which poses a human model first via its skeleton and then by applying fine-scale deformations.

\textbf{Learning and Networks:}
So far, our use of networks is limited to segmentation tasks, where we rely on segmentation masks for semi-automated tree branch labeling. 
Due to difficulties in getting sharp details from convolutional networks, the study of network-based segmentation of thin structures is still an active field in itself; 
there has been recent work on designing specialized multiscale architectures ~\cite{ronneberger:2015:unet,Lin_2017_CVPR,Qu:2018:STT:3240508.3240553} and also on incorporating perceptual losses~\cite{johnson2016perceptual} during network training~\cite{Mosinska_2018_CVPR}. 

\begin{figure}[t]
    \centering
    \ifdrawfigures
    \includegraphics[width=0.95\linewidth]{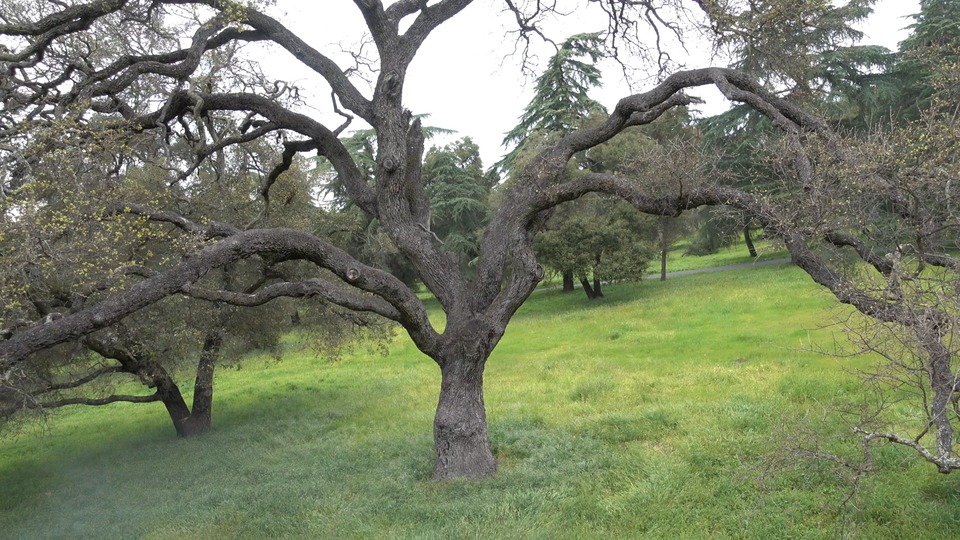}
    \llap{
        \makebox[0.95\linewidth][l]{
            \raisebox{0.05in}{
                \includegraphics[width=0.3\linewidth]{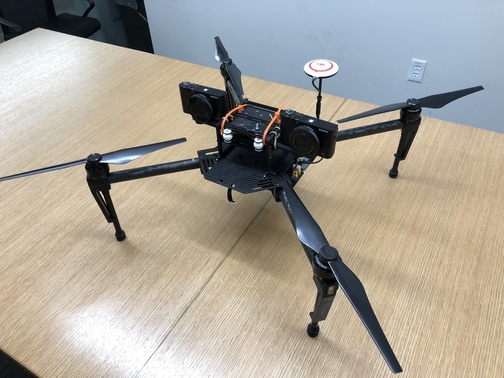}
            }
        }
    }
    \fi
    \caption{
        We target a California oak for reconstruction and simulation.
        (Inset) The drone and camera setup used to collect video data of the tree.
        \label{fig:overall_tree}}
    \ifusevspace
    \vspace{-.2in}
    \fi
\end{figure}

%%%%%%%%%%%%%%%%%%%%%%%%%%%%%%%%%%%%%%%%%%%%%%%%%%%%%%%%%%%%%%%%%%%%%%%%%%%%%%%
% Raw and Processed Data
%%%%%%%%%%%%%%%%%%%%%%%%%%%%%%%%%%%%%%%%%%%%%%%%%%%%%%%%%%%%%%%%%%%%%%%%%%%%%%%
\section{Raw and Processed Data}\label{sec:data}

As a case study, we select a California oak (\textit{quercus agrifolia}) as our subject for tree reconstruction and simulation (see Figure~\ref{fig:overall_tree}).
The mere size of this tree imposes a number of restrictions on our data capture: one has to deal with an outdoor, unconstrained environment, wind and branch motion will be an issue, it will be quite difficult to observe higher up portions of the tree especially at close proximities, there will be an immense number of occluded regions because of the large number of branches that one cannot see from any feasible viewpoint, etc.

\begin{figure*}[t]
    \centering
    \ifdrawfigures
    \includegraphics[width=.24\textwidth]{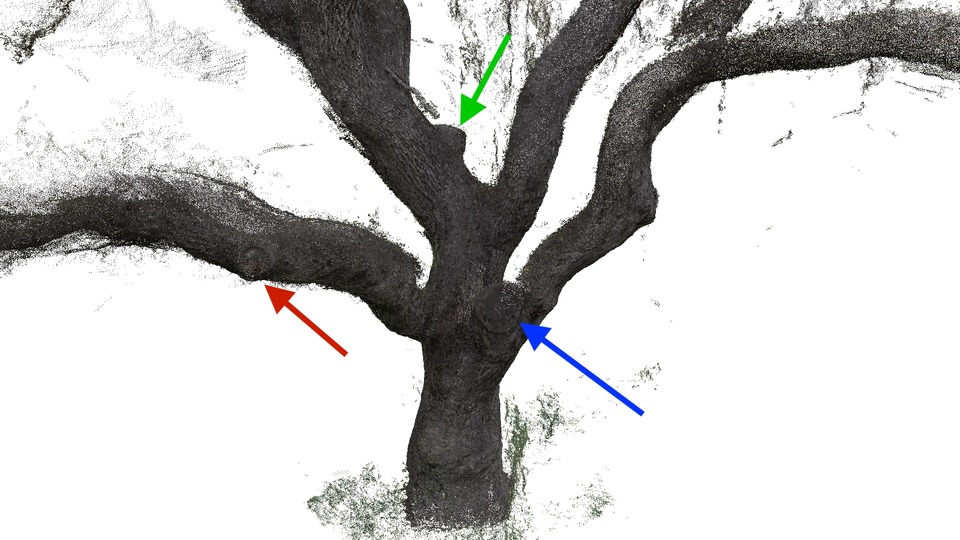}
    \includegraphics[width=.24\textwidth]{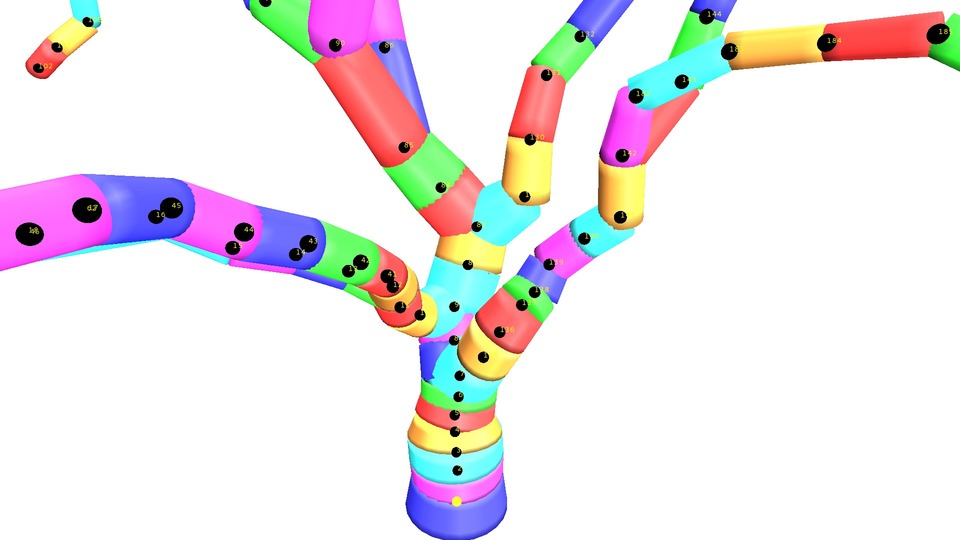}
    \includegraphics[width=.24\textwidth]{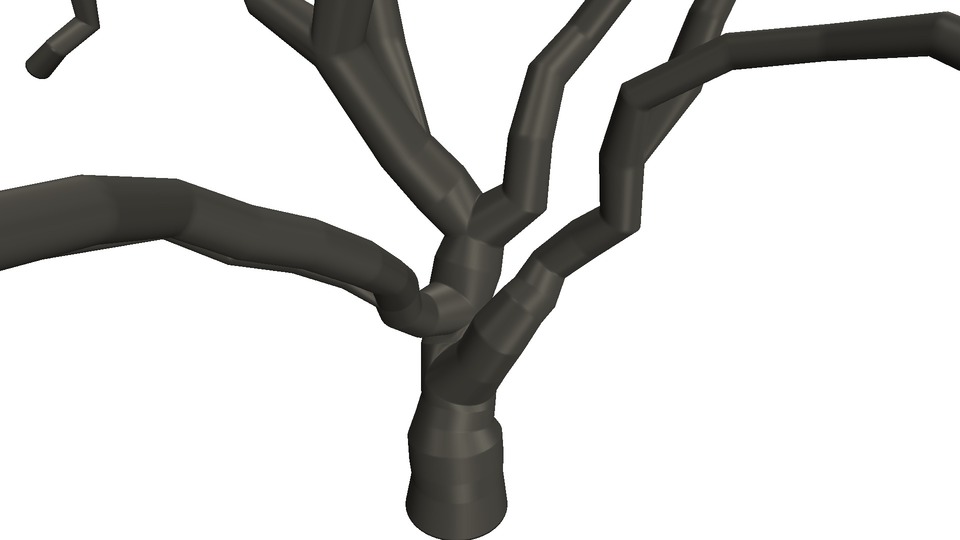}
    \includegraphics[width=.24\textwidth]{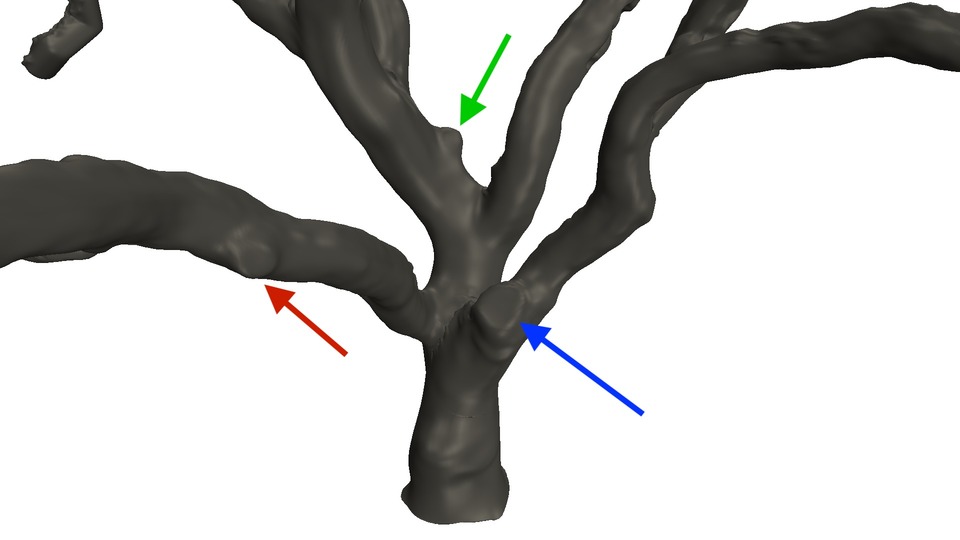}
    \fi
    \caption{
        Point cloud data (far left) is used as a guide for interactively placing generalized cylinders which are readily simulatable as articulated rigid bodies (middle left).
        The generalized cylinders are connected together into a skinned mesh (middle right), which is then perturbed based on the point cloud data to capture finer scale geometric details (far right).
        Notice the stumps that appear in the perturbed mesh reflecting the point cloud data (marked with arrows).
        \label{fig:coarse_model}}
    \ifusevspace
    \vspace{-.2in}
    \fi
\end{figure*}

In an outdoor setting, commodity structured light sensors that use infrared light (e.g.\ the Kinect) fail to produce reliable depth maps as their projected pattern is washed out by sunlight; thus, we opted to use standard RGB cameras.
Because we want good coverage of the tree, we cannot simply capture images from the ground; instead, we mounted our cameras on a quadcopter drone that was piloted around the tree.
The decision to use a drone introduces additional constraints: the cameras must be lightweight, the camera locations cannot be known \textit{a priori}, the drone creates its own air currents which can affect the tree's motion, etc.
Balancing the weight constraint with the benefits of using cameras with a global shutter and minimal distortion, we mounted a pair of Sony rx100 v cameras to a DJI Matrice 100 drone.
We calibrated the stereo offset between the cameras before flight, and during flight each camera records a video with $4$K resolution at $30$ fps.

Data captured in this manner is subject to a number of limitations.
Compression artifacts in the recorded videos may make features harder to track than when captured in a RAW format.
Because the drone must keep a safe distance from the tree, complete $360\degree$ coverage of a given branch is often infeasible.
This lack of coverage is compounded by occlusions caused by other branches and leaves (in seasons when the latter are present).
Furthermore, the fact that the tree may be swaying slightly in the wind even on a calm day violates the rigidity assumption upon which many multi-view reconstruction algorithms rely.
Since we know from the data collection phase that our data coverage will be incomplete, we will need to rely on procedural generation, inpainting, ``hallucinating" structure, etc.\ in order to complete the model.

After capturing the raw data, we augment it to begin to estimate the 3D structure of the environment.
We subsample the videos at a sparse $1$ or $2$ fps and use the Agisoft PhotoScan tool\footnote{Agisoft PhotoScan, \url{http://www.agisoft.com/}} to run structure from motion and multi-view stereo on those images, yielding a set of estimated camera frames and a dense point cloud.
We align cameras and point clouds from separate structure from motion problems by performing a rigid fit on a sparse set of control points.
This is a standard workflow also supported by open-source tools \cite{wu:2011:visualsfm,schonberger:2016:structure,moulon:2016:openmvg}.
Some cameras may be poorly aligned (or in some cases, so severely incorrect that they require manual correction).
Once the cameras are relatively close, one can utilize an inverse rendering approach like that of~\cite{loper:2014:opendr} adjusting the misaligned cameras' parameters relative to the point cloud.
In the case of more severely misaligned cameras, one may select correspondences between 3D points and points in the misaligned image and then find the camera's extrinsics by solving a perspective-\textit{n}-point problem~\cite{fischler:1981:random}.

%%%%%%%%%%%%%%%%%%%%%%%%%%%%%%%%%%%%%%%%%%%%%%%%%%%%%%%%%%%%%%%%%%%%%%%%%%%%%%%
% Building a Simulatable Tree
%%%%%%%%%%%%%%%%%%%%%%%%%%%%%%%%%%%%%%%%%%%%%%%%%%%%%%%%%%%%%%%%%%%%%%%%%%%%%%%
\section{Building a Simulatable Tree}\label{sec:sim}

Recent work enables the simulation of highly detailed trees modeled as articulated rigid bodies at real-time or interactive speeds ($9.5$k rigid bodies simulated at $86$ frames/sec, or $3$ million rigid bodies simulated at $2.3$ sec/frame) \cite{quigley:2018:real}.
The scalability of this method makes the simulation of rich, highly detailed geometric models of real-world trees feasible.
In order to apply such a method to our reconstruction, we need to create articulated rigid bodies with masses and inertia tensors connected via springs with stiffnesses and damping coefficients.
Unfortunately, point clouds, triangle soups with holes, and other similar 3D representations are not readily amenable to such an approach.
Commonly used techniques such as Poisson surface reconstruction \cite{kazhdan:2006:poisson} produce potentially disconnected meshes that do not respect the topology of the underlying tree, and are thus not well-suited for simulation.
In order to create a simulatable reconstructed tree, we make a strong prior assumption that the tree reconstruction consists of a number of generalized cylinders as underlying building blocks, appropriately skinned to provide smooth interconnections, and subsequently modified to provide the desired geometric detail.

We create the generalized cylinders interactively using the point cloud data obtained via multi-view stereo as a guide.
An initial cylinder is positioned at the base of the tree's trunk, then a second cylinder is attached to the first by a common endpoint, and so on, progressively ``growing" the generalized cylinder model of the tree.
Each cylinder endpoint may be connected to zero, one, or two subsequent cylinders to model a branch ending, curving, or bifurcating, respectively.
A radius is also specified for each endpoint of the generalized cylinders.
After approximating the trunk and branches using this generalized cylinder basis, the surfaces of the generalized cylinders are skinned together into a single contiguous triangle mesh representing the exterior surface of the tree.
Although this model only roughly captures the geometry using the radii of the generalized cylinders as estimates of the tree's cross-sectional thicknesses, the advantage of this representation is that the model has a topology consistent with the real tree.
The generalized cylinders are also readily simulated in order to drive deformations of the skinned mesh.
See Figure~\ref{fig:coarse_model}.

Although topologically accurate, the skinned generalized cylinders miss much of the rich geometric structure of the tree that is captured in part by the point cloud data.
Thus we augment the generalized cylinder representation in a manner informed by the point cloud.
For each vertex on the contiguous skinned mesh, we construct a cylindrical sampling region normal to the mesh; then, the average position of the point cloud points that fall within this sampling region is used to perturb the vertex in its normal direction, essentially creating a 2D height field with respect to the skinned mesh.
That is, the point cloud informs a displacement map~\cite{cook:1984:shade}.

Some vertices may have no point cloud data within their respective sampling regions; often the point cloud data only models one side of a branch, so only vertices on that side of the skin mesh are perturbed.
To avoid sharp discontinuities in the perturbed mesh, we solve Laplace's equation for the heightfield displacements of the vertices with empty sampling regions using the adequately perturbed heights as Dirichlet boundary conditions.
To obtain a visually desirable mesh, one can additionally utilize Laplacian smoothing (e.g.\ \cite{taubin:1995:signal,Desbrun:1999:implicitfairing}), vertex normal smoothing, Loop subdivision~\cite{loop:2001:subdivision}, etc., as well as point cloud subset selection interleaved with additional perturbations along the resulting vertex normal directions.
In fact, extending image inpainting ideas~\cite{bertalmio:2000:image} to the height fields on the two dimensional mesh surface (i.e.\ geometric inpainting) would likely give the best results, especially if informed from other areas of the tree where the geometry is more readily ascertained.

Finally, the mesh is textured by assigning each post-perturbation vertex the color of its nearest point cloud point.
Note that we do not use an average of the nearby data as this tends to wash out the texture details.
Here, image inpainting can also be used to fill in regions that have no point cloud data for textures.

%%%%%%%%%%%%%%%%%%%%%%%%%%%%%%%%%%%%%%%%%%%%%%%%%%%%%%%%%%%%%%%%%%%%%%%%%%%%%%%
% Medium-Scale Branches
%%%%%%%%%%%%%%%%%%%%%%%%%%%%%%%%%%%%%%%%%%%%%%%%%%%%%%%%%%%%%%%%%%%%%%%%%%%%%%%
\section{Medium Scale Branches}\label{sec:med_branches}

\begin{figure}[b]
    \ifusevspace
    \vspace{-.2in}
    \fi
    \centering
    \ifdrawfigures
    \twofigure{.235\textwidth}{.03in}{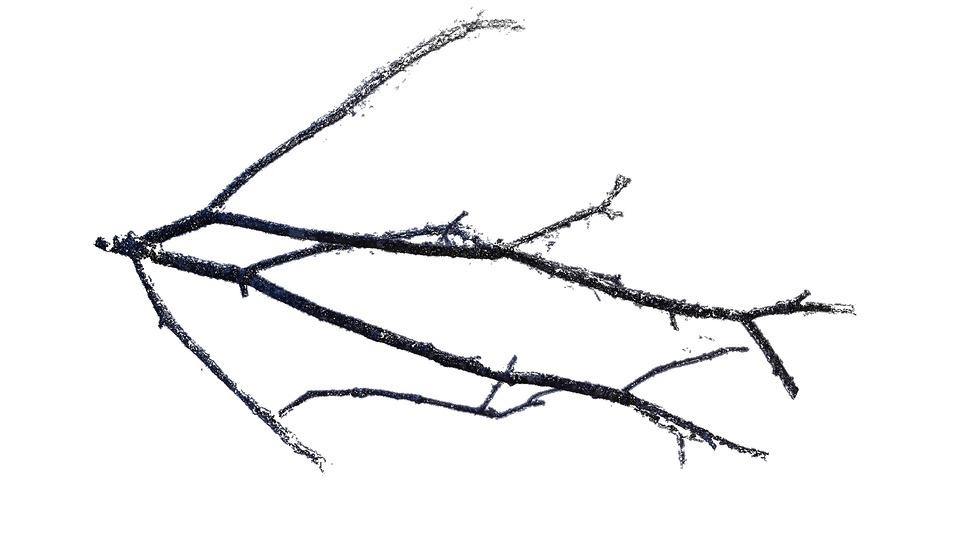}{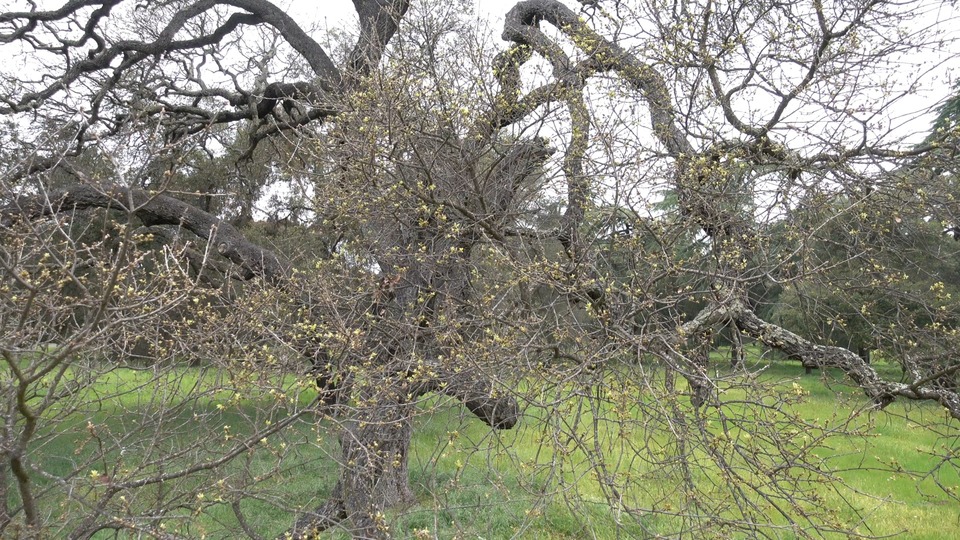}
    \fi
    \caption{
        (Left) We successfully reconstruct twig geometry using a traditional structure from motion and multi-view stereo pipeline under favorable conditions: complete coverage, indoor lighting, rigid geometry, etc.
        (Right) In the wild, difficult conditions preclude the ability for such reconstructions.
        \label{fig:vsfm_twig}}
\end{figure}

The aforementioned process fails on parts of the tree for which there is insufficient point cloud data (or no point cloud data at all).
Although traditional structure from motion is sufficient for recovering fine twig details under favorable conditions, the process of capturing data from a real tree accrues errors that necessitate a specialized approach (see Figure~\ref{fig:vsfm_twig}).
These errors may be attributed to many sources: some branches are heavily occluded by others, the drone cannot perform a full $360\degree$ sweep of most branches without other branches acting as obstacles, twig features are often only a few pixels wide when maintaining a safe distance between the drone and the tree, the tree may be nonrigidly deforming in the breeze even on a relatively calm day (or even due to the air currents generated by the drone itself), etc.
The net effect of these sources of error is that our approach for creating the trunk and thicker branches of the tree is insufficient for the tree's finer structures that are not well-resolved by the point cloud data.

\begin{figure}[b]
    \ifusevspace
    \vspace{-.25in}
    \fi
    \centering
    \ifdrawfigures
    \includegraphics[width=0.95\linewidth]{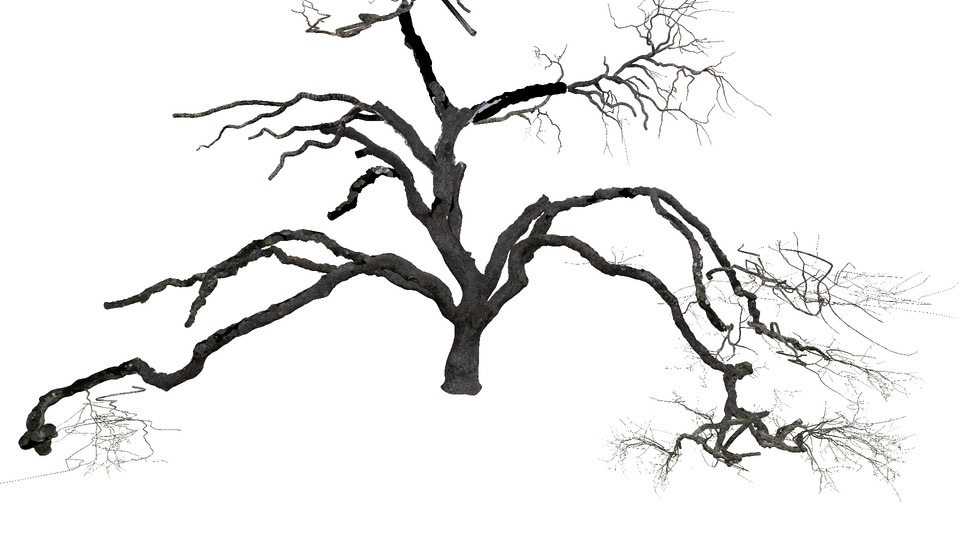}
    \includegraphics[width=0.95\linewidth]{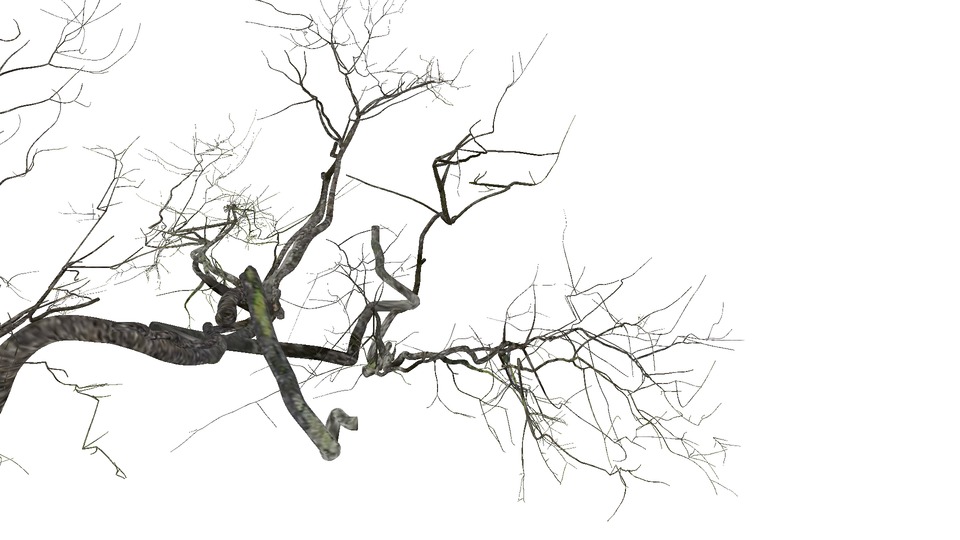}

    \fi
    \caption{
        (Top) Triangle meshes recovered from point cloud data (Section~\ref{sec:sim}) and image annotations (Section~\ref{sec:exp}).
        (Bottom) A close-up view of medium scale branches.
        \label{fig:skin_mesh}}
\end{figure}

Thus, we switch to an image-based approach for finer structures.
Our 3D generalized cylinder prior can be extended to 2D images by choosing projected radii and projected lengths of hypothetical 3D generalized cylinders.
These 2D projections of generalized cylinders can be extended back to three spatial dimensions using multiple images. %reprojected?
Whereas significant geometric detail on the thicker parts of the tree comes from geometric roughness on the skin of the cylinder as caused by knots, bark, etc.\ and is captured by our aforementioned perturbation process (see Section~\ref{sec:sim}), the most significant geometric detail on thinner branches is often simple bending of their centerline.
Thus, the fact that thinner branches lack adequate representation in the 3D point cloud is less consequential.

\begin{figure*}[t]
    \centering
    \ifdrawfigures
    \twofigure{.49\textwidth}{.05in}{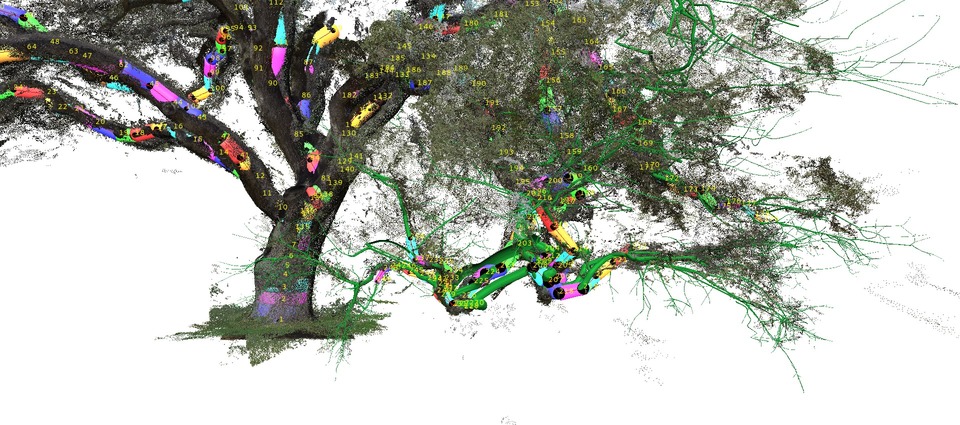}{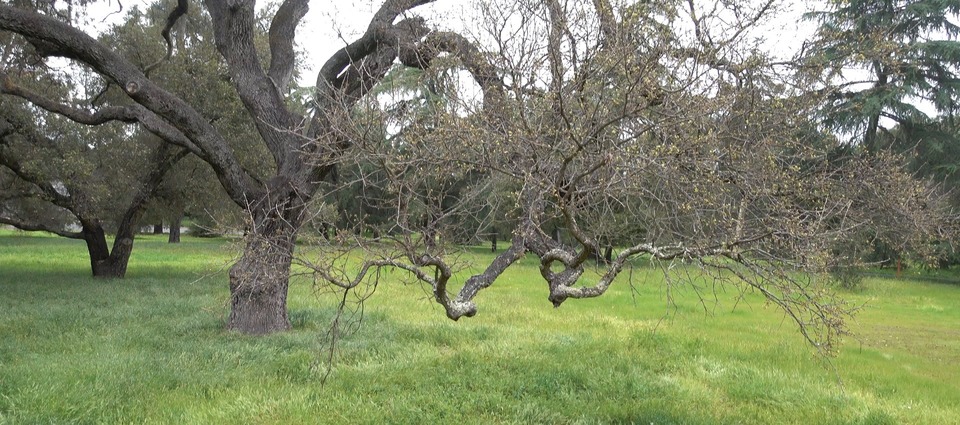}
    \fi
    \caption{
        (Left) Final simulatable geometry: articulated rigid bodies skinned and textured with the aid of the point cloud, along with thinner branch reconstructions with geometry and topology that follow the image data, all of which inform the motion of any reconstructed points and triangles that lack high enough fidelity reconstructions to form coherent structures.
        (Right) An RGB image of the actual tree from the same view.
        \label{fig:rgb_recon}}
    \ifusevspace
    \vspace{-.2in}
    \fi
\end{figure*}

We employ an image annotation approach (see Section~\ref{sec:annotation}) to obtain image space labelings of branch and twig curves and ``keypoints," or features that can be identified across multiple images.
After annotating a number of images, we use the annotation data to recover 3D structures by triangulating keypoint positions, connecting 3D keypoints to match the topology of image space curves, and estimating the tree thickness for each 3D point (see Section~\ref{sec:exp_med_branches}).
Then, we again create a contiguous skin mesh for these newly recovered 3D branches.
In order to boost our ability to capture geometric changes in the centerline, we use the 3D positions as control points for a b-spline curve rather than directly meshing the piecewise linear segments.
See Figure~\ref{fig:skin_mesh}.

Finally, we project texture information from the annotated images onto the skinned branches.
For each vertex on the skinned mesh, we estimate its corresponding position within each corresponding annotated curve by measuring its fractional length along the curve's medial axis and its fractional thickness measured by projecting the vertex's distance from its 3D segment onto a plane parallel to the current image plane.
For each such annotated curve we compute a quality estimate based on how close the corresponding camera is to the vertex and how closely aligned the vertex's surface normal is to the direction from the vertex to the annotated point.
Since averaging smears out texture information, we assign each vertex the color with the maximum quality score.

%%%%%%%%%%%%%%%%%%%%%%%%%%%%%%%%%%%%%%%%%%%%%%%%%%%%%%%%%%%%%%%%%%%%%%%%%%%%%%%
% Unresolved Structure
%%%%%%%%%%%%%%%%%%%%%%%%%%%%%%%%%%%%%%%%%%%%%%%%%%%%%%%%%%%%%%%%%%%%%%%%%%%%%%%
\section{Unresolved Structure}\label{sec:unresolved}

Because the image annotations depend on human labelers, many of the tree's branches and twigs remain unmodeled even as more images are progressively covered; the automated and semi-automated approaches considered in Section~\ref{sec:learning} can help with this.
In order to avoid discarding data, we additionally constrain the unstructured point cloud data obtained in Section~\ref{sec:data} to the nearest generalized cylinders of the reconstructed model so that the point cloud deforms as the tree's rigid bodies move during simulation.
This allows points from leaves, branches, and other structures that remain ``orphaned" even after all possible generalized cylinders are created to contribute to the virtual tree's appearance and motion.
See Figure~\ref{fig:rgb_recon}.

%%%%%%%%%%%%%%%%%%%%%%%%%%%%%%%%%%%%%%%%%%%%%%%%%%%%%%%%%%%%%%%%%%%%%%%%%%%%%%%
% Annotation and Learning
%%%%%%%%%%%%%%%%%%%%%%%%%%%%%%%%%%%%%%%%%%%%%%%%%%%%%%%%%%%%%%%%%%%%%%%%%%%%%%%
\section{Annotation and Learning}\label{sec:learning}

Annotating images is a challenging task for human labelers and automated methods alike.
Branches and twigs heavily occlude one another, connectivity can be difficult to infer, and the path of even a relatively large branch can often not be traced visually from a single view.
Thus it is desirable to augment the image data during annotation to aid human labelers.

One method for aiding the labeler is to automatically extract a ``flow field" of vectors tracking the anisotropy of the branches in image space (see Figure~\ref{fig:result_flow}).
The flow field is overlaid on the image in the annotation tool, and the labeler may select endpoints to be automatically connected using the projection-advection scheme discussed in Section~\ref{sec:learn_assist}.
Section~\ref{sec:learn_assist} also discusses how we generate the flow field itself, after first creating a segmentation mask.
Note that segmentation (i.e.\ discerning \emph{tree} or \emph{not tree} for each pixel in the image) is a simpler problem than annotation (i.e.\ discerning medial axes, topology, and thickness in image space).

Obtaining segmentation masks is straightforward under certain conditions, e.g.\ in areas where branches and twigs are clearly silhouetted against the grass or sky, but segmentation can be difficult in visually dense regions of an image.
Thus, we explore deep learning-based approaches for performing semantic segmentation on images from our dataset.
In particular, we use U-Net~\cite{ronneberger:2015:unet}, a state-of-the-art fully convolutional architecture for segmentation; the strength of this model lies in its many residual connections, which give the model the capacity to retain sharp edges despite its hourglass structure.
See Section~\ref{sec:deep_learn} for further discussion.

%%%%%%%%%%%%%%%%%%%%%%%%%%%%%%%%%%%%%%%%%%%%%%%%%%%%%%%%%%%%%%%%%%%%%%%%%%%%%%%
% Experiments
%%%%%%%%%%%%%%%%%%%%%%%%%%%%%%%%%%%%%%%%%%%%%%%%%%%%%%%%%%%%%%%%%%%%%%%%%%%%%%%
\section{Experiments}\label{sec:exp}

Since the approach to large scale structure discussed in Section~\ref{sec:sim} works well, we focus here on medium scale branches.

\subsection{Image Annotation}\label{sec:annotation}

\begin{figure}[t]
    \centering
    \ifdrawfigures
    \includegraphics[width=1.0\linewidth]{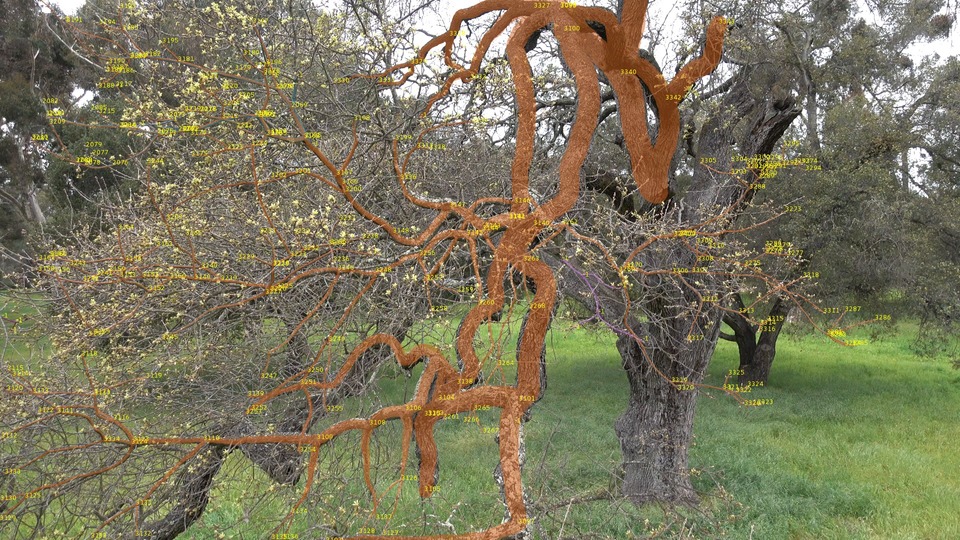}
    \fi
    \caption{Human labelers use our annotation tool to draw curves with positions, thicknesses, connectivities, and unique identifiers on images of the tree.\label{fig:annotation}}
    \ifusevspace
    \vspace{-.2in}
    \fi
\end{figure}

We present a human labeler with an interface for drawing piecewise linear curves on an overlay of a tree image.
User annotations consist of vertices with 2D positions in image space, per-vertex branch thicknesses, and edges connecting the vertices.
Degree-$1$ vertices are curve endpoints, degree-$2$ vertices lie on the interior of a curve, and degree-$3$ vertices exist where curves connect.
A subset of the annotated vertices are additionally given unique identifiers that are used to match common points between images; these will be referred to as ``keypoints" and are typically chosen as bifurcation points or points on the tree that are easy to identify in multiple images.
See Figure~\ref{fig:annotation}.

We take advantage of our estimated 3D knowledge of the tree's environment in order to aid human labelers and move towards automatic labeling.
After some annotations have been created, their corresponding 3D structures are generated and projected back into each image, providing rough visual cues for annotating additional images.
Additionally, since we capture stereo information, we augment our labeling interface to be aware of stereo pairs: users annotate one image, copy those annotations to the stereo image, and translate the curve endpoints along their corresponding epipolar lines to the correct location in the stereo image.
This curve translation constrained to epipolar lines (with additional unconstrained translation if necessary to account for error) is much less time consuming than labeling the stereo image from scratch.

Human labelers often identify matching branches and twigs across images by performing human optical flow, toggling between adjacent frames of the source video and using the parallax effect to determine branch connectivity.
This practice is an obvious candidate for automation, e.g.\ by annotating an initial frame then automatically carrying the annotated curves through subsequent frames via optical flow.
Unfortunately, the features of interest are often extremely small and thin and the image data contains compression artifacts, making automatic optical flow approaches quite difficult.
However, it is our hope that in future work the same tools that aid human labelers can be applied to automatic approaches making them more effective for image annotation.

\subsection{Deep Learning}\label{sec:deep_learn}
In order to generate flow fields for assisting the human labeler as discussed in Section~\ref{sec:learning}, we first obtain semantic segmentations of \emph{tree} and \emph{not tree} using a deep learning approach.
To train a network for semantic segmentation, we generate a training dataset by rasterizing the image annotations as binary segmentation masks of the labeled branches.
From these 4K masks, we then generate a dataset of $512\times512$ crops containing more than 4000 images.
The crop centers are guaranteed to be at least $50$ pixels away from one another, and each crop is guaranteed to correspond to a segmentation mask containing both binary values.
The segmentation problem on the raw 4K images must work on image patches with distinctly different characteristics: the more straightforward case of branches silhouetted against the grass, and the more complex case of highly dense branch regions.
Therefore, we split the image patches into two sets via $k$-means clustering, and train two different models to segment the two different cases.
For the same number of training epochs, our two-model approach yields qualitatively better results than the single-model approach.

\begin{figure}[t]
    \centering
    \ifdrawfigures
    \includegraphics[width=1.0\linewidth]{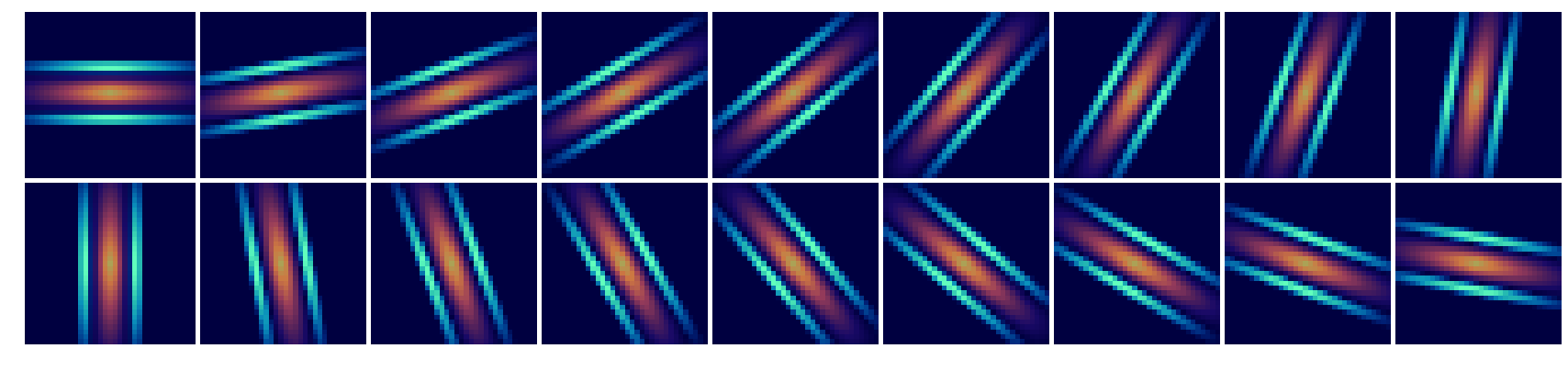}
    \fi
    \caption{
        A set of anisotropic kernels is used to obtrain directional activations in segmentation masks for both perceptual loss and flow field generation.
        \label{fig:kernels}}
    \ifusevspace
    \vspace{-.2in}
    \fi
\end{figure}

Instead of directly using the standard binary cross entropy loss, the sparseness and incompleteness of our data led us to use a weighted variant, in order to penalize false negatives more than false positives.
As a further step to induce smoothness and sparsity in our results, we introduce a second order regularizer through the L2 difference of the output and ground truth masks' gradients. 
We also experiment with an auxiliary loss similar to the VGG perceptual loss described in \cite{Mosinska_2018_CVPR}, but instead of using arbitrary feature layers of a pretrained network, we look at the L1 difference of hand-crafted multiscale directional activation maps.
These activation maps are produced by convolving the segmentation mask with a series of Gabor filter-esque~\cite{jain:1991:gaborsegmentation} feature kernels $\{k(\theta,r,\sigma): \mathbb{R}^2\rightarrow [0,\ldots,N]^2\}$, where each kernel is scale-aware and piecewise sinusoidal (see Figure~\ref{fig:kernels}).
A given kernel $k(\theta,r,\sigma)$ detects branches that are at an angle $\theta$ and have thicknesses within the interval $[r,\sigma r]$. 
For our experiments, we generate 18 kernels spaced $10$ degrees apart and use $N=35$, $r=4$, and $\sigma=1.8$.

\begin{figure}[t]
    \centering
    \ifdrawfigures
    \twofigure{0.48\linewidth}{0.03in}{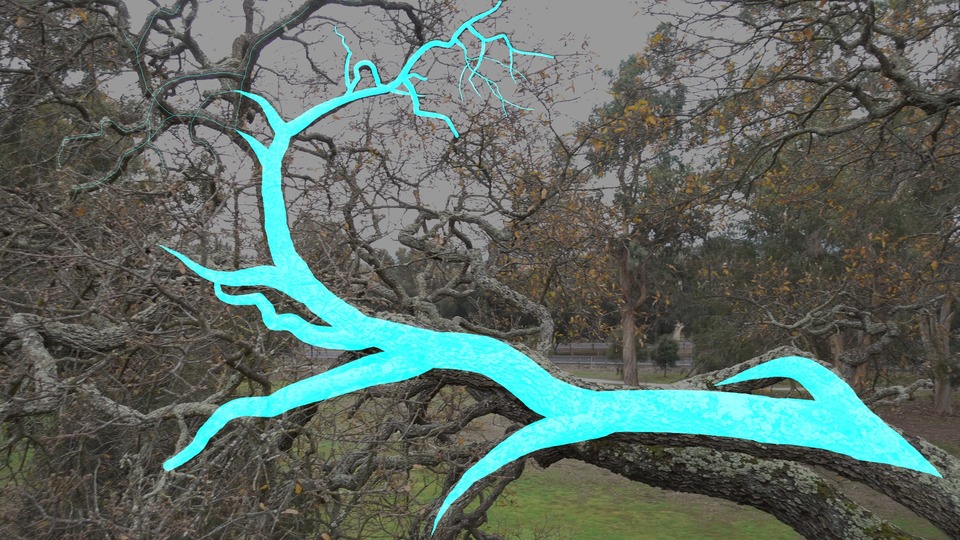}{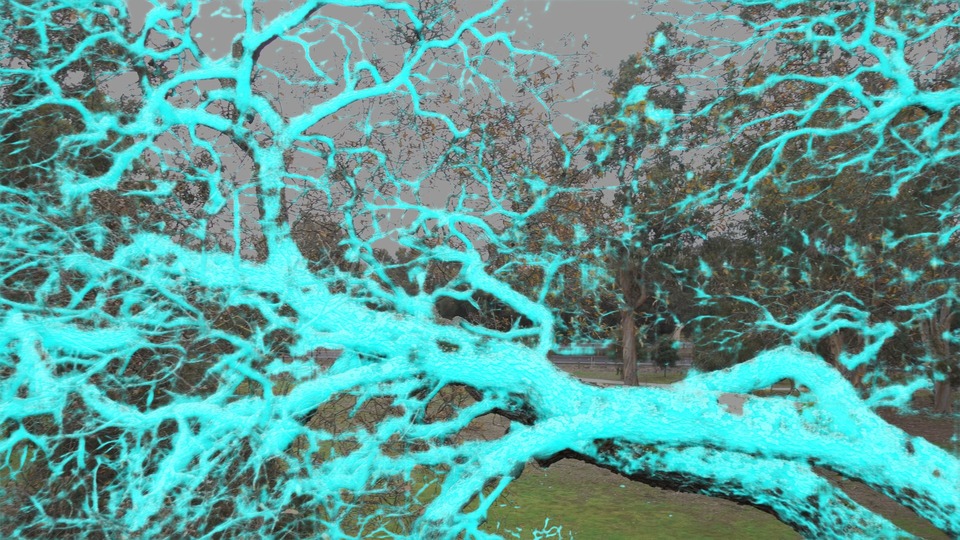}
    \vspace{0.015in}
    \twofigure{0.48\linewidth}{0.03in}{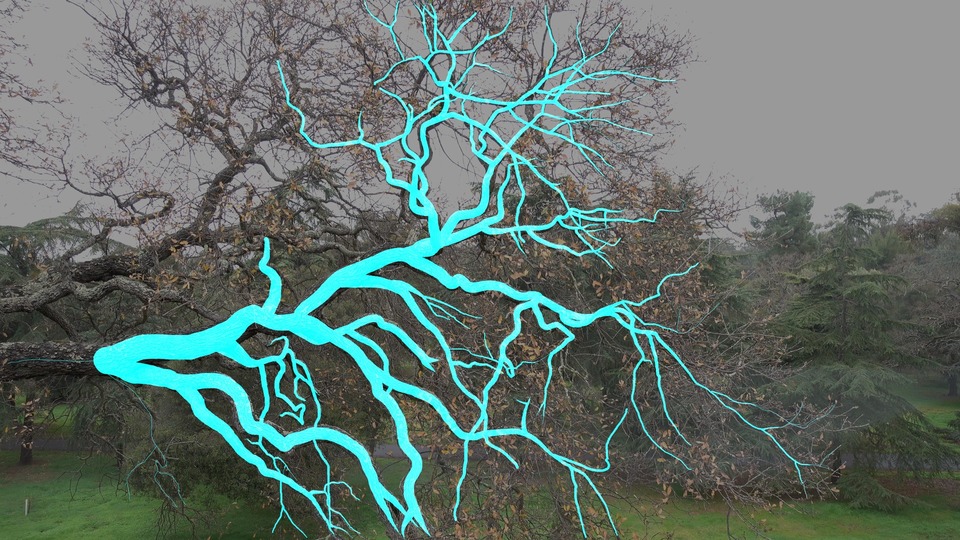}{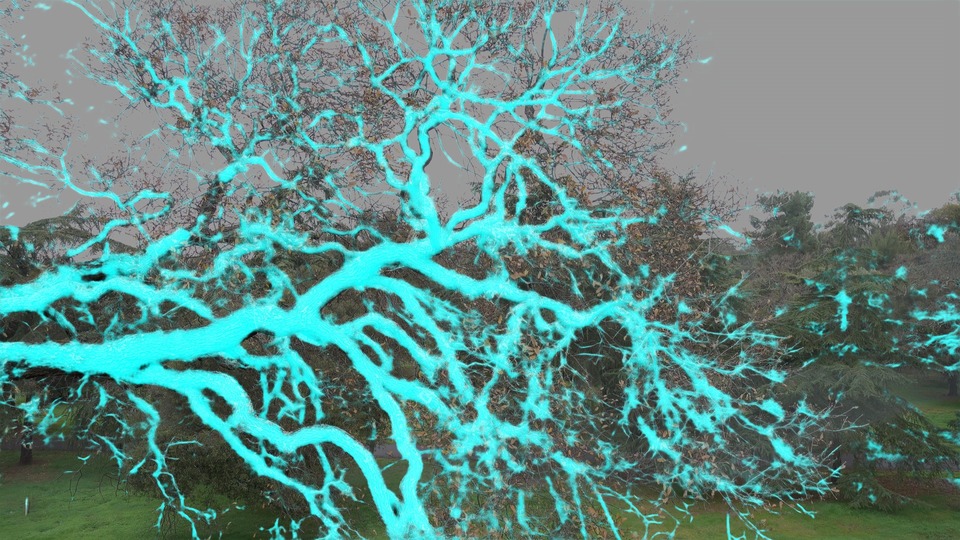}
    \fi
    \caption{
        (Left) Image masks generated from image annotations and used as training data.
        (Right) Outputs of the segmentation network.
        \label{fig:dataset}}
    \ifusevspace
    \vspace{-.2in}
    \fi
\end{figure}

Figure~\ref{fig:dataset} illustrates two annotated images used in training and the corresponding learned semantic segmentations.
Note that areas of the semantic segmentation that are not part of the labeled annotation may correspond to true branches or may be erroneous; for the time being a human must still choose which pieces of the semantic segmentation to use in adding further annotations.

\subsection{Learning-Assisted Annotation}\label{sec:learn_assist}

\begin{figure}[b]
    \ifusevspace
    \vspace{-.2in}
    \fi
    \centering
    \ifdrawfigures
    \vbox{
        \hbox{
            \vbox{
                \hbox{\includegraphics[width=0.325\linewidth]{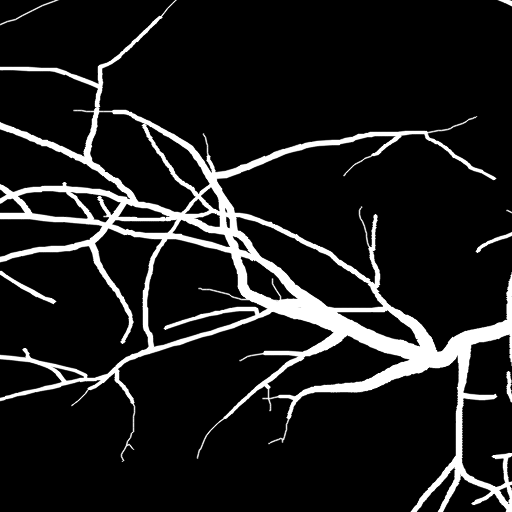}}
                \hbox{\includegraphics[width=0.325\linewidth]{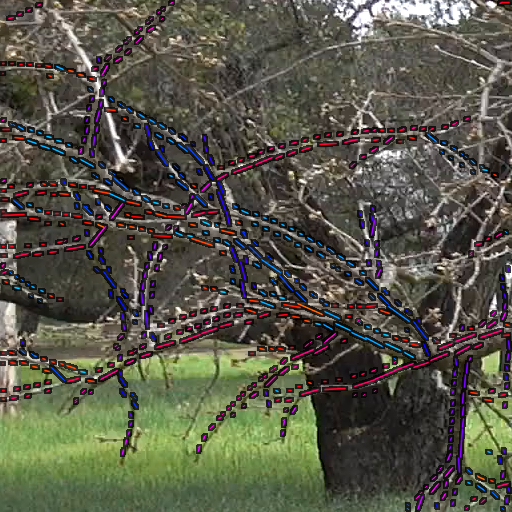}}
            }
            \includegraphics[width=0.65\linewidth]{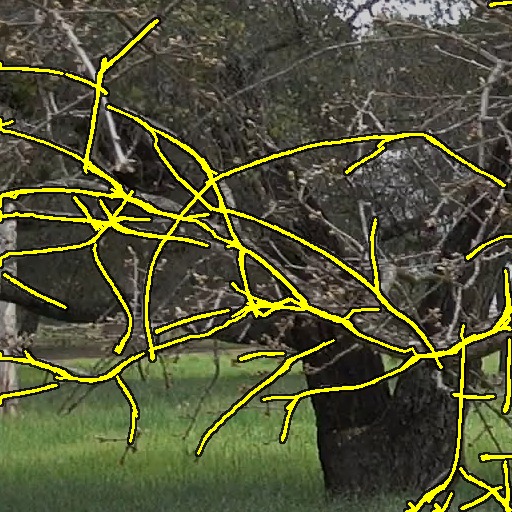}
        }
    }
    \fi
    \caption{
        (Top left) A ground truth mask of the tree taken by flattening the image annotation data into a simple binary mask.
        (Bottom left) A visualization of flow directions estimated by applying directional filters to the ground truth mask.
        (Right) Medial axes of the tree branches estimated from the flow field.
        \label{fig:groundtruth_flow}}
\end{figure}

To generate a flow field, we create directional activation maps as in Section~\ref{sec:deep_learn} again using the kernels from Figure~\ref{fig:kernels}, then perform a clustering step on the resulting per-pixel histograms of gradients~\cite{dalal2005histograms} to obtain flow vectors.
Each pixel labeled as \emph{tree} with sufficient confidence is assigned one or more principal directions; pixels with more than one direction are potentially branching points. 
We find the principal directions by detecting clusters in each pixel's activation weights; for each cluster, we take the sum of all relevant directional slopes weighted by their corresponding activation values.

\begin{figure}[b]
    \ifusevspace
    \vspace{-.2in}
    \fi
    \centering
    \ifdrawfigures
    \vbox{
        \hbox{
            \vbox{
                \hbox{\includegraphics[width=0.325\linewidth]{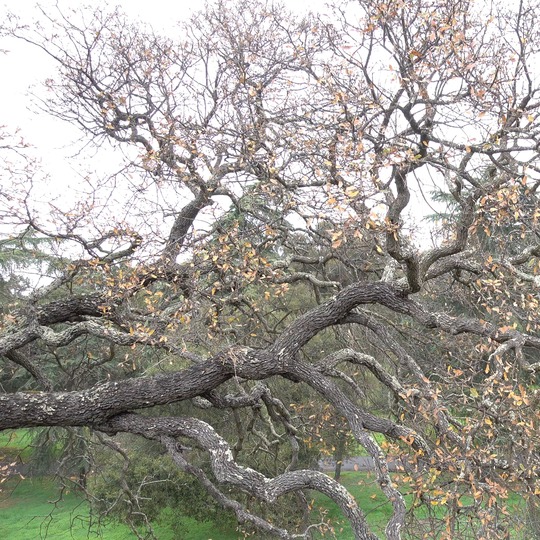}}
                \hbox{\includegraphics[width=0.325\linewidth]{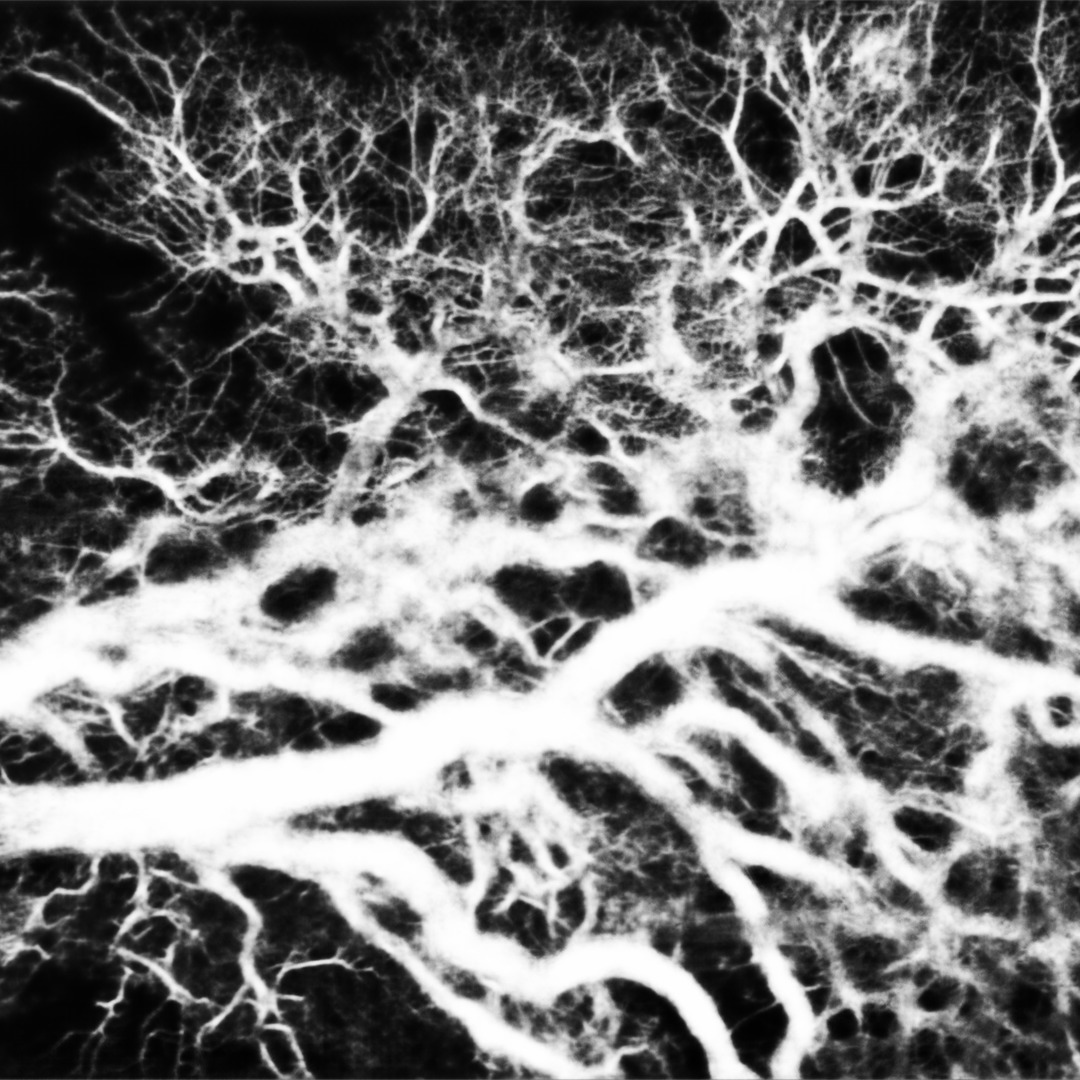}}
            }
            \includegraphics[width=0.65\linewidth]{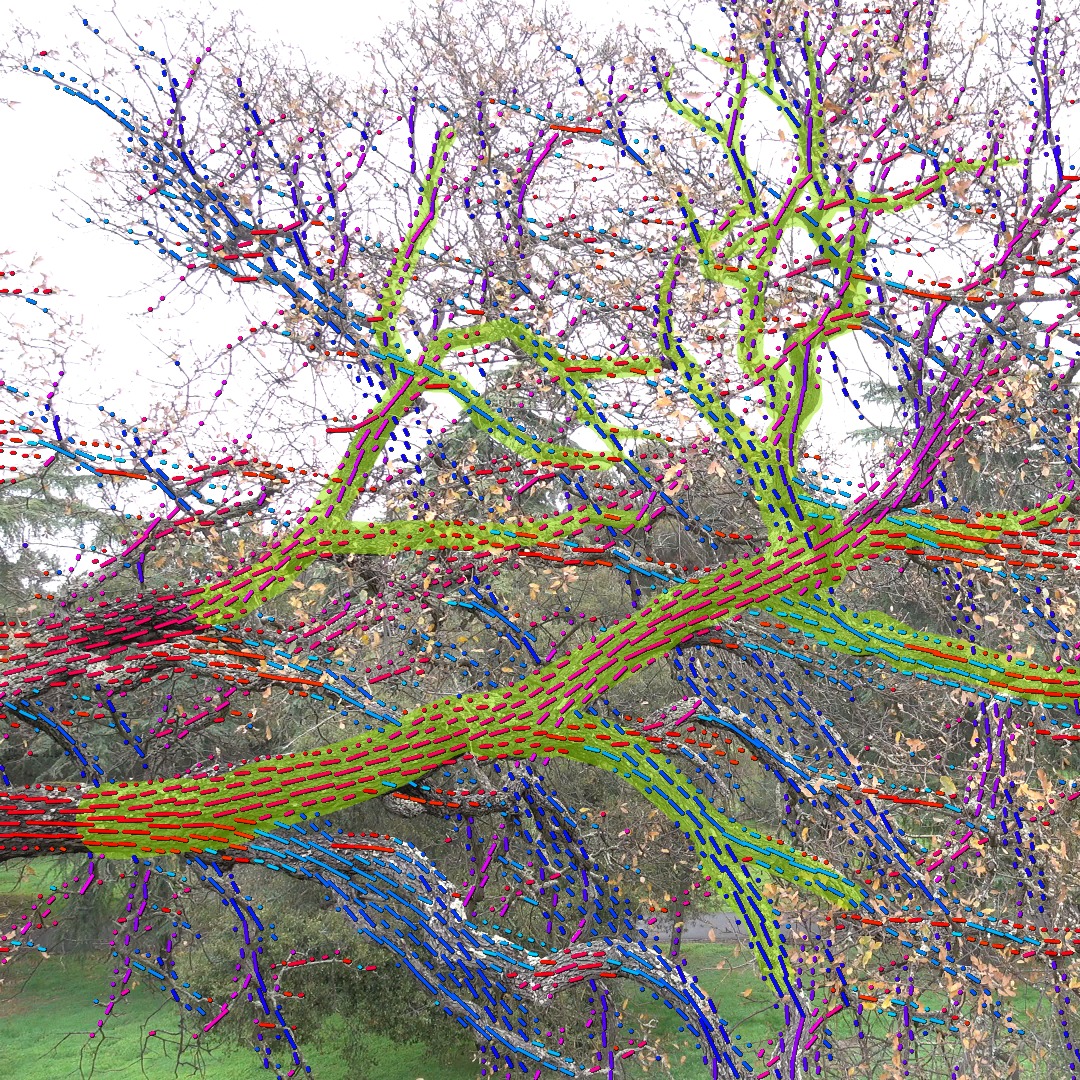}
        }
    }
    \fi
    \caption{
        The trained network infers a segmentation mask (bottom left) from an input image (top left).
        We then estimate a flow field (right) by applying anisotropic filters to the segmentation mask.
        A labeler can specify endpoints between which a medial axis and thickness values are automatically estimated (right, in green).
        \label{fig:result_flow}}
\end{figure}

Having generated a flow field of sparse image space vectors, we trace approximate medial axes through the image via an alternating projection-advection scheme.
From a given point on a branch, we estimate the thickness of the branch by examining the surrounding flow field and project the point to the estimated center of the branch.
We then advect the point through the flow field and repeat this process.
In areas with multiple directional activations (e.g.\ at branch crossings or bifurcations), our advection scheme prefers the direction that deviates least from the previous direction.
See Appendix~\ref{sec:proj_adv} for further details.
By applying this strategy to flow fields generated from ground truth image segmentations, we are able to recover visually plausible medial axes (see Figure~\ref{fig:groundtruth_flow}).
However, medial axes automatically extracted from images without ground truth labels are error prone.
Thus, we overlay the flow field on the annotation interface and rely on the human labeler.
The labeler may select curve endpoints in areas where the flow field is visually plausible, and these endpoints are used to guide the medial axis generation.
See Figure~\ref{fig:result_flow} for an example flow field generated from the learned segmentation mask and the supplemental video for a demonstration of semi-automated medial axis generation.

\subsection{Recovering Medium Scale Branches}\label{sec:exp_med_branches}

Given a set of image annotations and camera extrinsics obtained via structure from motion and stereo calibration, we first construct piecewise linear branches in 3D.
We triangulate keypoints that have been labeled in multiple images, obtaining 3D positions by solving for the point that minimizes the sum of squared distances to the rays originating at each camera's optical center and passing through the camera's annotated keypoint.
We then transfer the topology of the annotations to the 3D model by connecting each pair of 3D keypoints with a line segment if a curve exists between the corresponding keypoint pair in any image annotation.

Next, we subdivide and perturb the linear segments connecting the 3D keypoints to match the curvature of the annotated data.
Each segment between two keypoints is subdivided by introducing additional vertices evenly spaced along the length of the segment.
For each newly introduced vertex, we consider the point that is the same fractional length along the image-space curve between the corresponding annotated keypoints in each image for which such a curve exists.
We trace rays through these intra-curve points to triangulate the position of each new vertex in the same way that we triangulated the original keypoints.

\begin{figure}[b]
    \ifusevspace
    \vspace{-.2in}
    \fi
    \centering
    \ifdrawfigures
    \twofigure{.23\textwidth}{.03in}{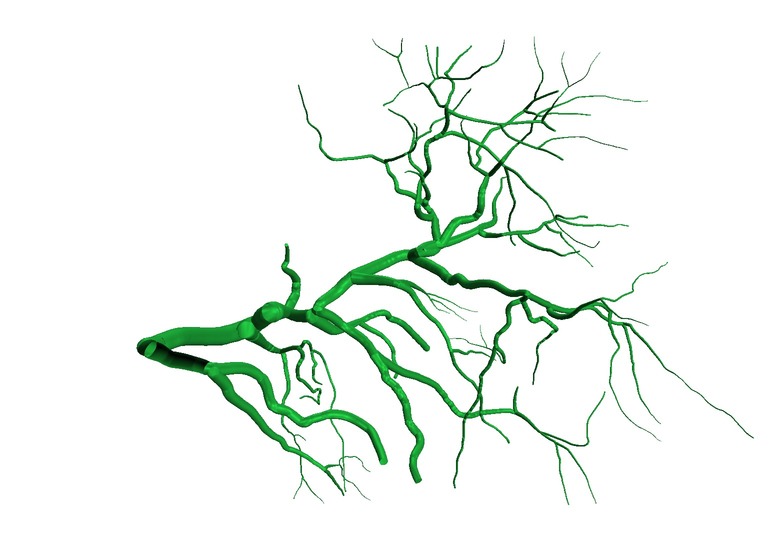}{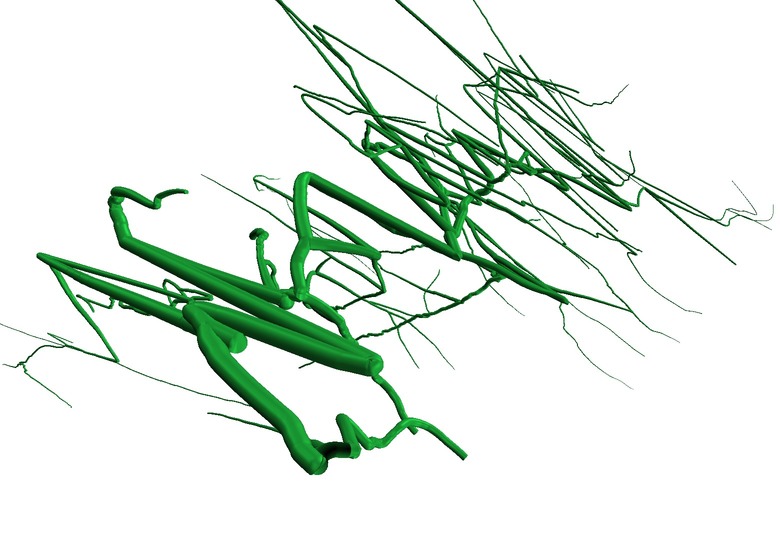}
    \twofigure{.23\textwidth}{.03in}{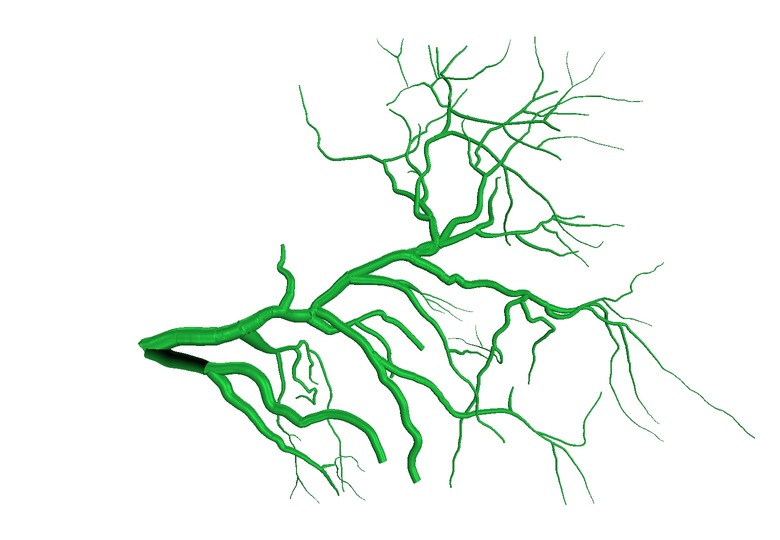}{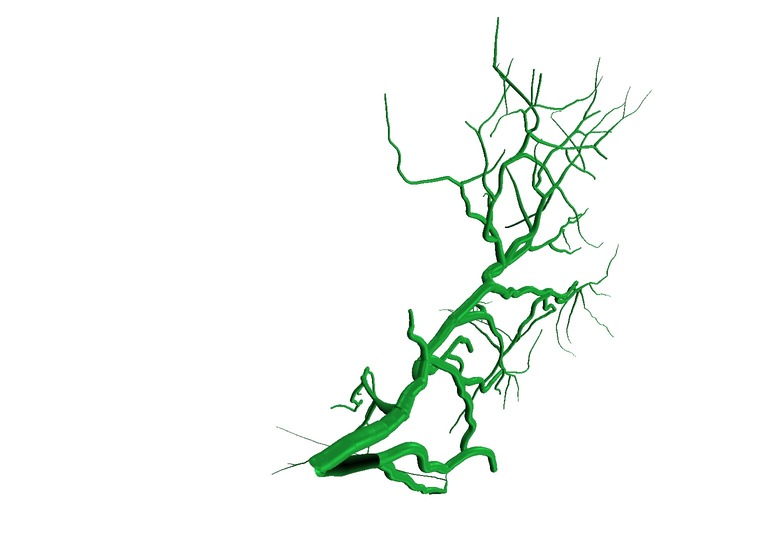}
    \fi
    \caption{
        Branches labeled from a stereo pair of cameras are visually plausible from the perspective of those cameras (top left), but they can exhibit severe error when viewed from a different angle (top right).
        By clamping these branch positions, one can achieve a virtually identical projection to the original cameras (bottom left) while maintaining a nondegenerate albeit ``flattened" appearance from a different angle (bottom right).
        \label{fig:clamping}}
\end{figure}

Finally, we estimate the thickness of each 3D vertex beginning with the 3D keypoints.
We estimate the world space thickness of each keypoint by considering the corresponding thickness in all annotated camera frames.
For each camera in which the keypoint is labeled, we estimate world space thickness using similar triangles, then average these estimates to get the final thickness value.
We then set the branch thickness of each of the vertices obtained through subdivision simply by interpolating between the thicknesses of the keypoints at either end of the 3D curve.
Using this strategy, we recover a set of 3D positions with local cross-sectional thicknesses connected by edges, which is equivalent to the generalized cylinder representation employed in Section~\ref{sec:sim}.

The human users of our annotation tools encounter the traditional trade-off of stereo vision: it is easy to identify common features in images with a small baseline, but these features triangulate poorly exhibiting potentially extreme variance in the look-at directions of the corresponding cameras.
Conversely, cameras whose look-at directions are close to orthogonal yield more stable triangulations, but common features between such images are more difficult to identify.
One heuristic approach is to label each keypoint three times: twice in similar images and once from a more diverse viewpoint.
However, it may be the case that some branches are only labeled in two images with a small baseline (e.g.\ a stereo pair).
In this case, we propose a clamping strategy based on the topological prior of the tree.
Designating a ``root" vertex of a subtree for such annotations, we triangulate the annotated keypoints as usual obtaining noisy positions in the look-at directions of the stereo cameras.
We then march from the root vertex to the leaf vertices.
For each vertex $p$ with location $p_x$, we consider each outboard child vertex $c$ with location $c_x$.
For each camera in which the point $c$ is labeled, we consider the intersection of the ray from the location of $c$'s annotation to $c_x$ with the plane parallel to the image plane that contains $p_x$; let $c_x'$ be the intersection point.
We then clamp the location of $c$ between $c_x'$ and the original location $c_x$ based on a user-specified fraction.
This process is repeated for each camera in which $c$ is annotated, and we obtain the final location for $c$ by averaging the clamped location from each camera.
See Figure~\ref{fig:clamping}.

%%%%%%%%%%%%%%%%%%%%%%%%%%%%%%%%%%%%%%%%%%%%%%%%%%%%%%%%%%%%%%%%%%%%%%%%%%%%%%%
% Conclusion and Future Work
%%%%%%%%%%%%%%%%%%%%%%%%%%%%%%%%%%%%%%%%%%%%%%%%%%%%%%%%%%%%%%%%%%%%%%%%%%%%%%%
\section{Conclusion and Future Work}\label{sec:conclusion}

\begin{figure}[t]
    \centering
    \ifdrawfigures
    \twofigure{.235\textwidth}{.03in}{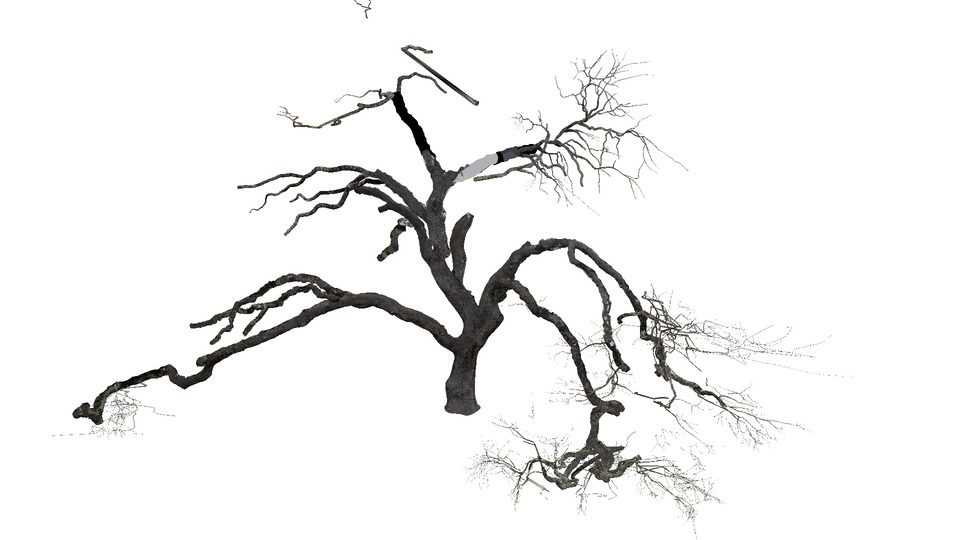}{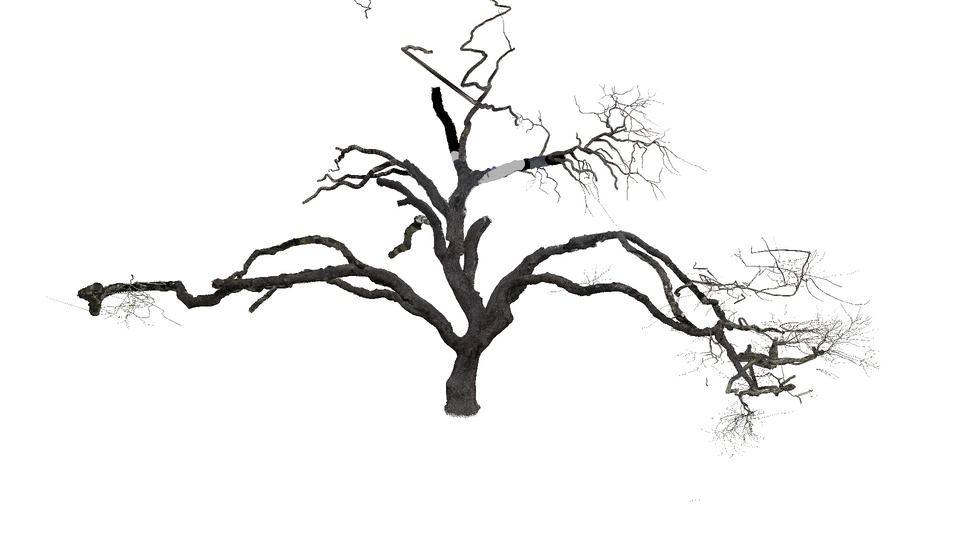}
    \fi
    \caption{
        The tree model is deformed from its rest pose (left) to an exaggerated pose (right) via simulation.
        \label{fig:sim}}
    \ifusevspace
    \vspace{-.2in}
    \fi
\end{figure}

We presented an end-to-end pipeline for reconstructing a 3D model of a botanical tree from RGB image data.
Our reconstructed model may be readily simulated to create motion in response to external forces, e.g.\ to model the tree blowing in the wind (see Figure~\ref{fig:sim}).
We use generalized cylinders to initialize an articulated rigid body system, noting that one could subdivide these primitives as desired for additional bending degrees of freedom, or decrease their resolution for faster performance on mobile devices.
The simulated bodies drive the motion of the textured triangulated surfaces and/or the point cloud data as desired.

Although we presented one set of strategies to go all the way from the raw data to a simulatable mesh, it is our hope that various researchers will choose to apply their expertise to and improve upon various stages of this pipeline, yielding progressively better results.
In particular, the rich topological information in the annotations has great potential for additional deep learning applications, particularly for topology extraction~\cite{ventura:2017:iterative,mattyus:2017:deeproadmapper,xue:2018:seeing} and 2D to 3D topology generation~\cite{estrada:2015:tree}.

\section*{Acknowledgments}
Research supported in part by ONR N000014-13-1-0346, ONR N00014-17-1-2174, ARL AHPCRC W911NF-07-0027, and generous gifts from Amazon and Toyota.
In addition, we would like to thank both Reza and Behzad at ONR for supporting our efforts into computer vision and machine learning, as well as Michael Black for many interesting suggestions especially regarding drones.
E.Q. was supported in part by an NDSEGF and Y.Z. was supported in part by a Stanford School of Engineering Fellowship.
E.Q. would also like to thank Michael Bao and David Hyde for contributing cameras, Pilot AI Labs for contributing drones, and Richard Heru, Katherine Liu, Martin Mbuthia, and Riley Wilson for helping with image annotation.

\appendix

\renewcommand{\thesection}{\Alph{section}}
\section*{Appendices}

\section{Video Syncing}
We collect data using a pair of drone-mounted cameras.
In order to synchronize the two video feeds, we calculate a sequence of the L1 distances between adjacent video frames for each camera as a rough estimate of how much motion occurred between frames.
We then use the maximum cross correlation of these sequences to find the integer frame offset that best aligns the two videos.
Note that since the videos are recorded at $30$ fps, even the optimal offset could leave as much as a 1/60 second temporal error, in addition to the other sources of error mentioned in Section~\ref{sec:med_branches}.

\section{Camera Calibration}
For each structure from motion problem, we withhold images from camera 2 and obtain estimates of camera 1's intrinsics as well as extrinsics for individual image samples.
Then, we hold the camera intrinsics constant and solve for the stereo offset between cameras 1 and 2 using a standard checkerboard pattern~\cite{zhang:2000:flexible}.
The scale of the stereo transform relative to the structure from motion scene is unknown, but this scale may be found using image correspondences or even estimated visually.

\section{Semantic Segmentation}
The branches and twigs in the image data are often either clustered together or silhouetted against the ground.
Hypothesizing that two models trained on identical network architectures might perform better than a single monolithic model, we divided the training data into a `brown' set and a `green' set using $k$-means clustering approximately capturing the `clustered' and `silhouetted' arrangements of twigs, respectively.
Because color saturation is a defining feature in green vs. brown, we chose to cluster the image crops based on the saturation channel in HSV color space.
However, using the average saturation over the entire image crop is not discriminatory enough due to the high amounts of brown that are present even in crops that contain branches silhouetted against grass; the median saturation value is also not sufficiently robust to varying amounts of grass in the background.
Thus, for each image crop we took the $30$\textsuperscript{th} and $90$\textsuperscript{th} percentile of per-pixel saturations as a feature vector.
We performed $k$-means clustering on these 2D features to split the image crops into two groups.
Finally, we trained two models with identical U-Net network architectures, each with image crops of the clusters.

\begin{figure*}[h]
    \centering
    \vbox{
        \includegraphics[width=0.85\textwidth]{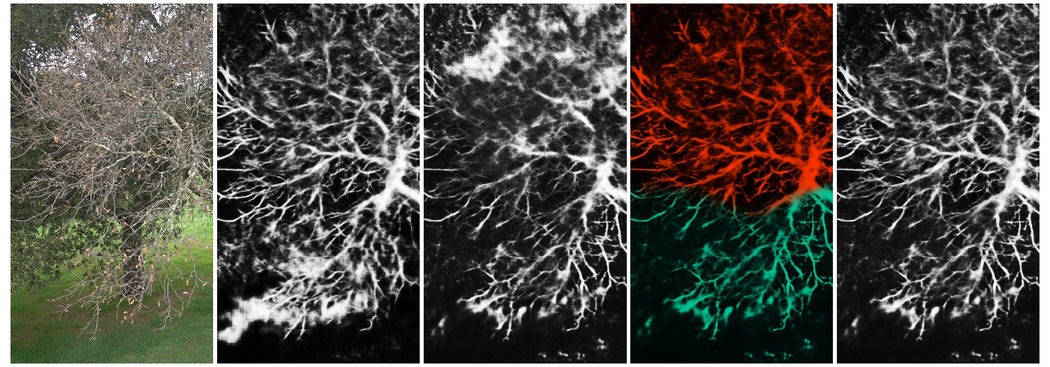}
     }
    \caption{
        From left to right: original image, output of model trained on cluster 1 (`brown'), output of model trained on cluster 2 (`green'), visualization of contributions from each model in the composited output (orange for cluster 1 and green for cluster 2), and the final composited segmentation.
        \label{fig:segmentation_clustering}}
\end{figure*}

\begin{figure*}[h]
    \centering
    \includegraphics[width=1.0\textwidth]{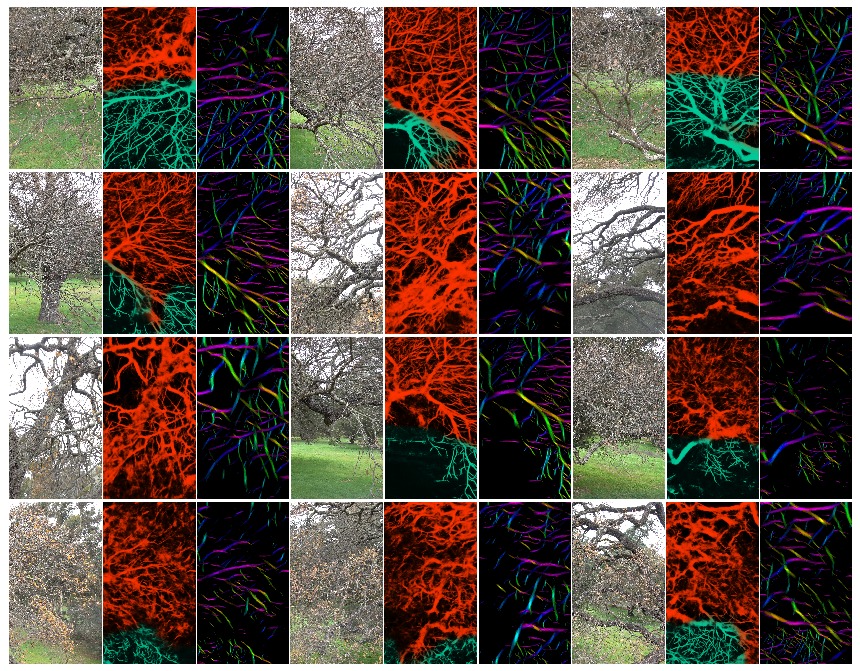}
    \caption{
        Examples of network-predicted segmentations and subsequent flow field extractions using images not seen during network training. 
            \label{fig:segmentation_examples}}
\end{figure*}

To segment an image, we pass the image through both models to obtain two candidate segmentation masks.
We then composite the candidate segmentations by taking a per-pixel weighted sum.
To set the per-pixel weights, we compute a feature vector of the $30$\textsuperscript{th} and $90$\textsuperscript{th} percentile saturation values for a small neighborhood around each pixel; then, we calculate the feature vector's distance to each of the two cluster centers previously obtained in the $k$-means clustering step and use those distances to assign the corresponding pixel a weight between $0$ and $1$.
See Figures~\ref{fig:segmentation_clustering} and~\ref{fig:segmentation_examples} for examples.

\begin{figure*}[h]
    \centering
    \vbox{
        \includegraphics[width=0.85\textwidth]{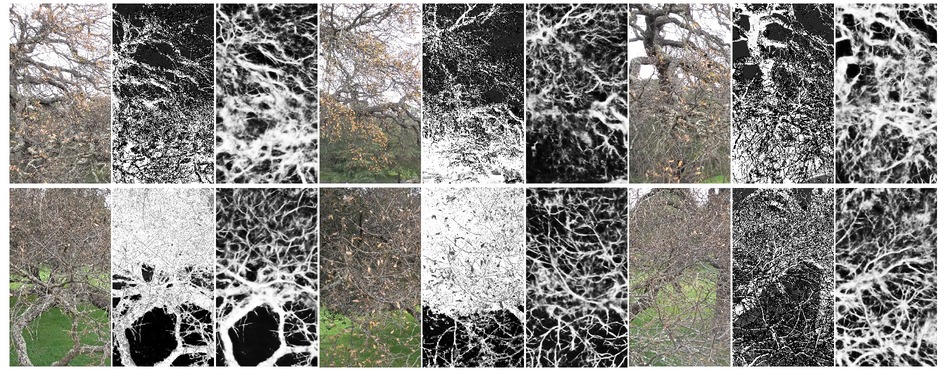}
    }
    \caption{
        Examples of $k$-means clustering versus our trained segmentation network.
        From left to right for each image triplet: input image, $k$-means result, network output.
        For this visualization, we performed $k$-means clustering with 2 clusters in L$^*$a$^*$b$^*$ color space, which has a Euclidean distance that closely corresponds to color differences perceived by humans.
        \label{fig:segmentation_kmeans}}
\end{figure*}

Note that while our experiments with $k$-means clustering on training data for network models were reasonably successful, experiments with using clustering algorithms directly on images to perform segmentation were distinctively less so, as the color-based segmentation methods had trouble segmenting out tangled branches that are not clearly silhouetted against the grass or the sky.
See Figure~\ref{fig:segmentation_kmeans} for several examples.

\section{Flow Field Extraction} 

\begin{figure*}[h]
    \centering
    \vbox{
        \includegraphics[width=0.9\linewidth]{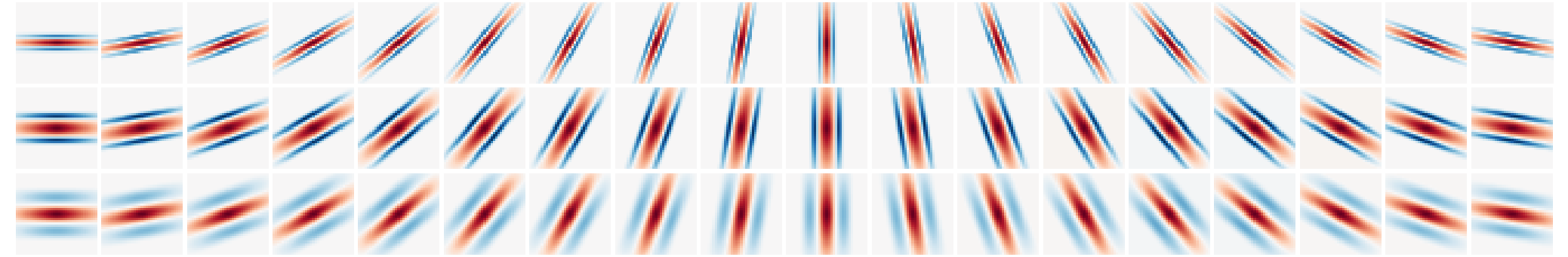}
    }
    \caption{
        Heatmap visualizations of $35\times35$ kernels, where red indicates positive weights and blue indicates negative weights.
        The middle row shows kernels with $r=3.8$, $\sigma=0.8$, and $\theta$ increasing in $10\degree$ increments; these are the values we use in practice.
        To illustrate the effect of the kernel parameters, the top and bottom rows depict kernels with $r=2.0$, $\sigma=0.8$ and $r=3.8$, $\sigma=2.0$, respectively.
        \label{fig:flow_kernels}}
\end{figure*}

Our handcrafted scale-aware directional kernels are piecewise sinusoidal with a linear falloff, and are designed to have a maximum activation when convolved with an image patch that contains a line passing through the patch center with some specified thickness $r$ and orientation $\theta$.
A falloff threshold $\sigma$ controls the maximum detectable thickness; convolving any line thicker than $(1+\sigma)r$ with one of the kernels will result in an activation value of zero.
See Figure~\ref{fig:flow_kernels}.
The formulation of the kernel $k: S \rightarrow \mathbb{R}$ defined on the discrete pixel grid $S=\{-17,-16,\ldots, 16, 17\}^2$ is
\begin{equation*}\begin{split}
k(p\textrm{ ; }\theta,r,\sigma)&=
\begin{cases}w_{p;r}\cos(\frac{d_{p;\theta,r}}{2}\pi) & \textrm{if }d_{p;\theta,r}<1\\
-w_{p;r}\frac{1}{\sigma}\sin(\frac{(d_{p;\theta,r}-1)}{\sigma}\pi)&\textrm{if }1<d_{p;\theta,r}<1+\sigma\\
0&\textrm{if } 1+\sigma<d_{p;\theta,r}\end{cases}\\
\textrm{where } 
 w_{p;r}&=1-\frac{1}{r}||p||\textrm{ , }\\
 v_\theta&=\begin{pmatrix}\cos\theta\\\sin\theta\end{pmatrix}\textrm{ , }\\
 d_{p;\theta,r}&=\frac{1}{r}||p-(p\cdot v_\theta)v_\theta ||.
 \end{split}\end{equation*}

\begin{figure*}[h]
    \centering
        \begin{subfigure}{0.22\textwidth}
            \includegraphics[width=1.0\textwidth]{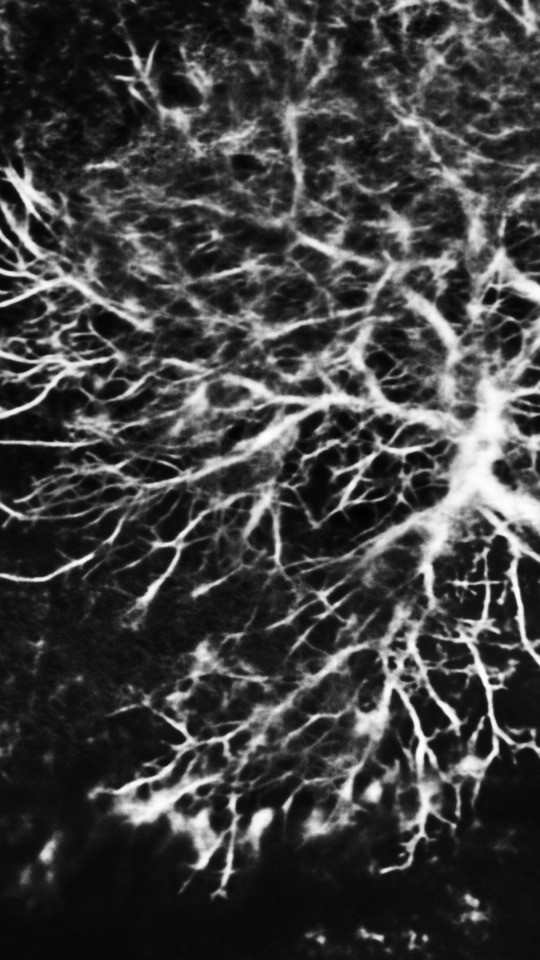}
            \caption{}
        \end{subfigure}
        \begin{subfigure}{0.22\textwidth}
            \includegraphics[width=1.0\textwidth]{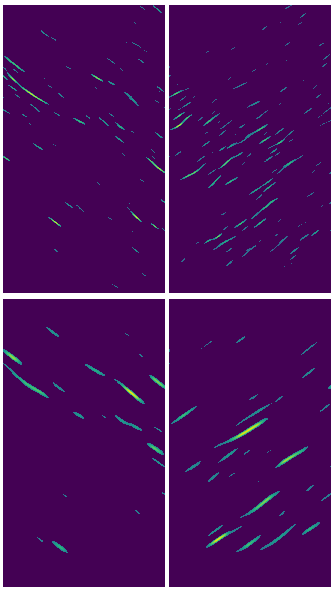}
            \caption{\label{subfig:filters}}
        \end{subfigure}
        \begin{subfigure}{0.22\textwidth}
            \includegraphics[width=1.0\textwidth]{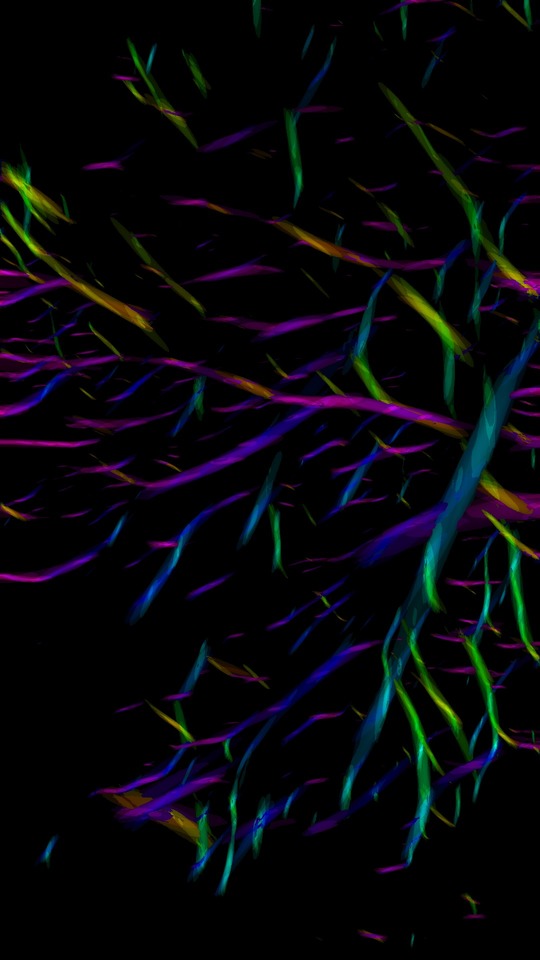}
            \caption{}
        \end{subfigure}
        \begin{subfigure}{0.22\textwidth}
            \includegraphics[width=1.0\textwidth]{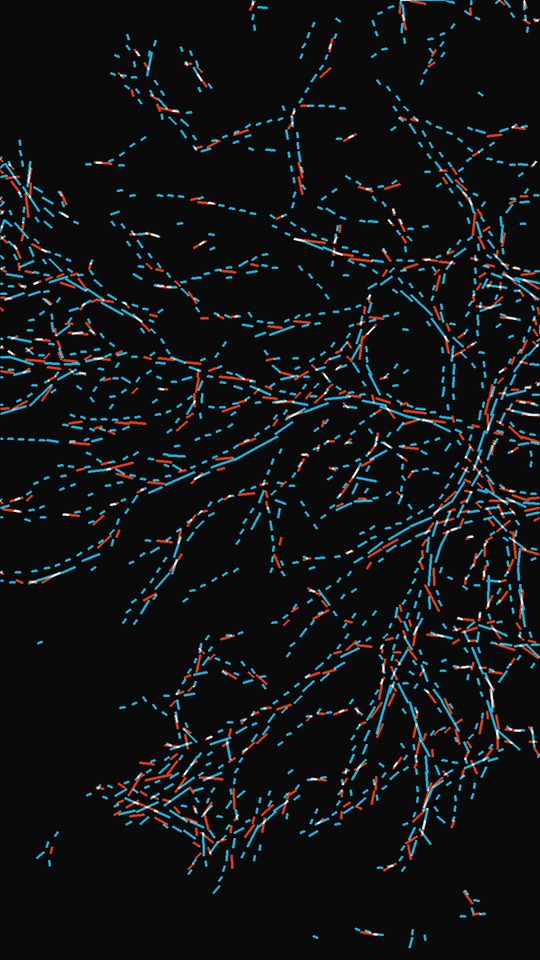}
            \caption{}
        \end{subfigure}
    \caption{
        From left to right: input segmentation mask, a selected subset of multiscale directional activation maps, an HSV visualization of primary flow directions (in which hue represents angle and value represents magnitude), and a flow field visualization of primary and secondary flow directions (visualized in blue and orange, respectively).
        \label{fig:segmentation_and_flow}}
\end{figure*}

\begin{figure*}[ht]
    \centering
    \vbox{
        \begin{subfigure}{1.0\linewidth}
            \centering
            \makebox[0.65\linewidth][c]{
                \includegraphics[width=0.2\linewidth]{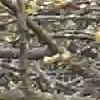}
                \includegraphics[width=0.2\linewidth]{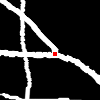}
                \includegraphics[width=0.2\linewidth]{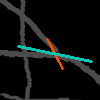}
            }
            \caption{
                An image crop (left), its corresponding ground truth segmentation mask with a pixel of interest colored in red (middle), and the two directions detected at the pixel of interest during flow field generation (right).
                \label{subfig:filter_images}
                }
        \end{subfigure}
        \begin{subfigure}{1.0\linewidth}
            \centering
            \includegraphics[width=0.65\linewidth]{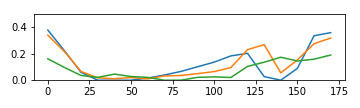}
            \caption{
                Plot of filter activation vs filter angle (in degrees) for the pixel of interest in (a).
                Each colored line corresponds to the activations at that pixel's location for a different image scale.
                \label{subfig:filter_activation}
                }
        \end{subfigure}
        \begin{subfigure}{1.0\linewidth}
            \centering
            \includegraphics[width=0.65\linewidth]{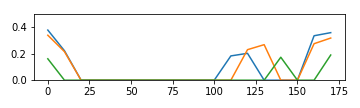}
            \caption{
                We discard activations from (b) that are below a user-specified threshold.
                \label{subfig:filter_threshold}
                }
        \end{subfigure}
        \begin{subfigure}{1.0\linewidth}
            \centering
            \includegraphics[width=0.65\linewidth]{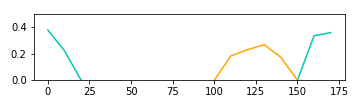}
            \caption{
                We reduce the activations from (c) to a single activation map by taking the maximum value over all scales.
                These values are grouped into clusters bounded by local minima (here blue and orange represent two clusters).
                Note that a cluster may wrap around from $170\degree$ to $0\degree$ since the filters are bidirectional.
                \label{subfig:filter_reduce}
                }
        \end{subfigure}
        \begin{subfigure}{1.0\linewidth}
            \centering
            \includegraphics[width=0.65\linewidth]{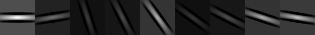}
            \caption{
                Kernels corresponding to nonzero values in the final activations from (d).
                \label{subfig:filter_kernels}
                }
        \end{subfigure}
    }
    \caption{
        An illustration of how filter activations are grouped into clusters as part of the flow field generation process.
        \label{fig:activation_blocks}}
\end{figure*}

We use a bank of $18$ filters, with $\theta$ ranging from $0\degree$ to $170\degree$ in $10\degree$ increments.
Instead of varying $r$ and $\sigma$, for more efficient computation we use fixed, empirically chosen values for $r$ and $\sigma$ and convolve these filters with a number of rescaled instances of the input image.
These convolutions yield a set of directional activations (see Figure~\ref{subfig:filters} and Figure~\ref{subfig:filter_activation}).

To extract flow field vectors from the directional activations, we first determine whether a meaningful nonzero flow direction exists for each pixel.
First, we filter out noise by zeroing out all activation values that are below an empirically chosen threshold (see Figure~\ref{subfig:filter_threshold}).
Next, for each pixel and for each $\theta$ sample, we reduce the activation maps for different scales to a single map by taking the maximum activation across all scales.
This yields a per-pixel histogram of activation values as a function of $\theta$.
See Figure~\ref{subfig:filter_reduce}.

Pixels with all zero activations and pixels with more than half their activations nonzero are ignored; the latter case is for filtering out pixels corresponding to roughly isotropic `blobs' in the segmentation mask.
We determine primary directions for the remaining pixels by clustering their activation values.
Observing that an elongated structure in the segmentation mask may cause large activations in multiple kernels with similar $\theta$ values, we sort the directional activations by $\theta$ and find blocks of activations caused by adjacent $\theta$ samples bounded by local minima in the distribution.
See Figures~\ref{subfig:filter_reduce} and~\ref{subfig:filter_kernels}.
For a given activation block $b$ with per-angle activations $a_{\theta}$, we use the weighted sum of the directions $\sum_{\theta\in b}{a_{\theta}\big(\begin{smallmatrix}\cos\theta\\\sin\theta\end{smallmatrix}\big)}$ as a single vector representing the block.
After finding all such blocks and computing the corresponding vectors, we designate the vectors with the largest magnitudes as the primary flow directions for the corresponding pixel.
In practice, selecting the top two vectors as primary flow directions has proven sufficient so far.
See Figure~\ref{subfig:filter_images} and Figure~\ref{fig:segmentation_and_flow}.

\section{Projection-Advection Scheme}\label{sec:proj_adv}
For a point $p$ with nonzero flow $\boldsymbol{n}$, we consider the line $L$ perpendicular to $\boldsymbol{n}$ that passes through $p$.
We then find the line segment $\overline{AB}$ on $L$ which includes $p$ and for which every point on the segment has a principal flow direction within $15$ degrees of $\boldsymbol{n}$.
The center point $(A+B)/2$ of the line segment, referred to as $p'$, is now the projection of $p$ onto the medial axis, and the length $\vert\vert\overline{AB}\vert\vert$ is the thickness of the branch at point $p'$.
$p'$ is added to the set of medial axis vertices, and we next consider the flow field at point $p'$. 
If the flow field at $p'$ is nonzero, we choose the principle direction $\boldsymbol{n'}$ that is closest to $\boldsymbol{n}$.
$p$ is then updated to be $p'+t\boldsymbol{n'}$, where $t$ is a user-specified step size, and the projection-advection scheme is repeated until we find a zero flow field at $p'$, i.e.\ where a nonzero $\boldsymbol{n'}$ does not exist.

Note that when estimating a medial axis within our annotation tool we additionally have labeler-specified curve endpoints $p_0$ and $p_1$.
To account for this we add an additional attraction force, picking the advection direction $\boldsymbol{n'}$ as $w \boldsymbol{n}_f + (1-w) \boldsymbol{n}_d$, where $\boldsymbol{n}_f$ is the normalized flow field direction, $\boldsymbol{n}_d$ is the normalized direction from $p'$ to $p_1$, and $w=\frac{||p'-p_1||}{||p_0-p_1||}$.

\section{Recovering Medium Scale Branches}

\begin{figure*}[h]
    \centering
    \twofigure{0.48\textwidth}{0.2in}{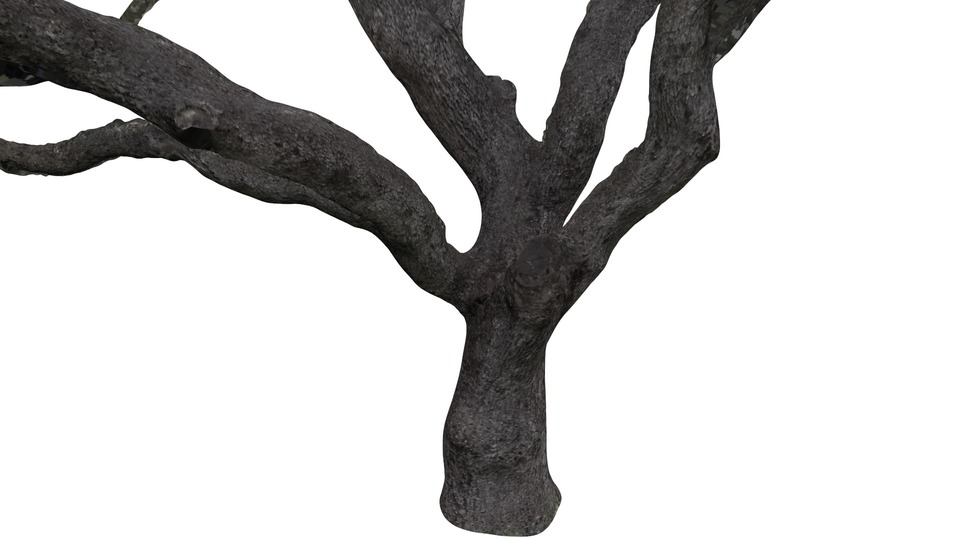}{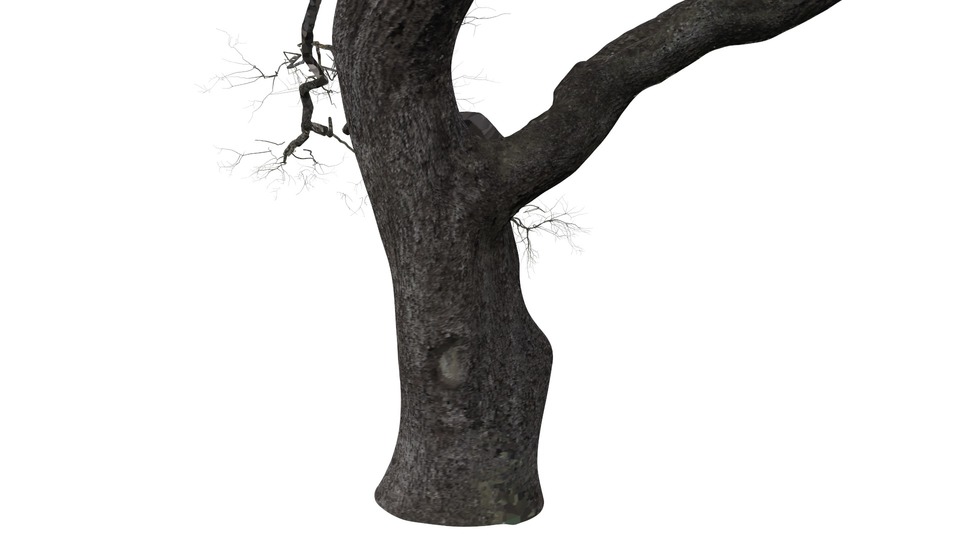}
    \twofigure{0.48\textwidth}{0.2in}{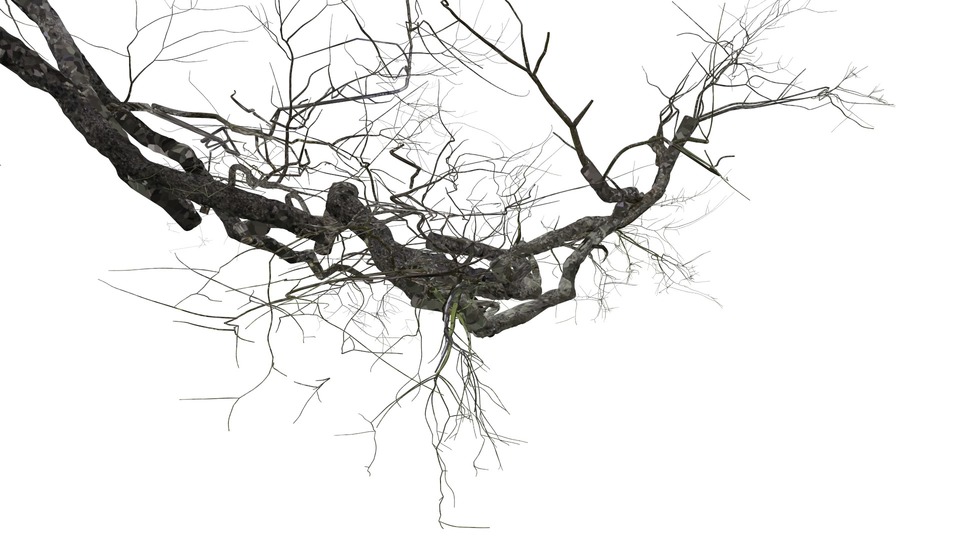}{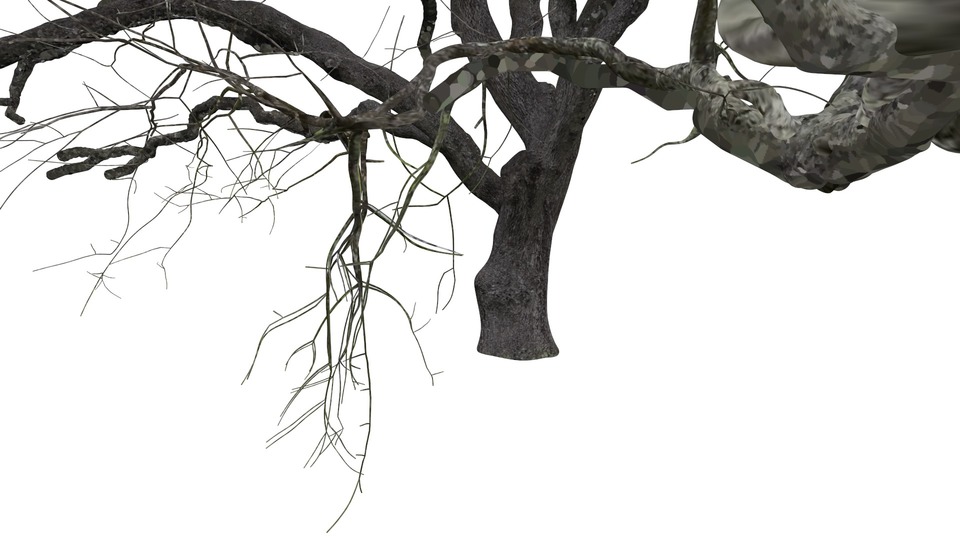}
    \twofigure{0.48\textwidth}{0.2in}{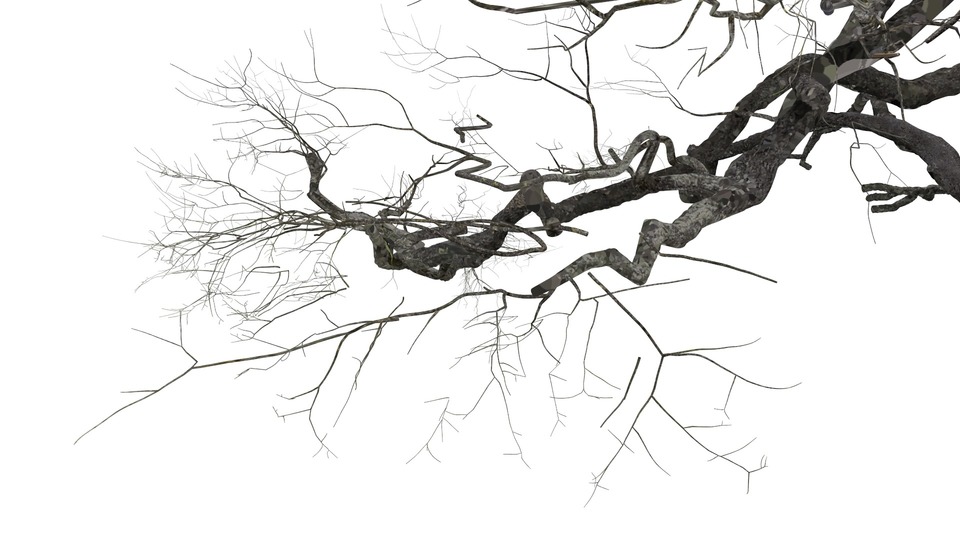}{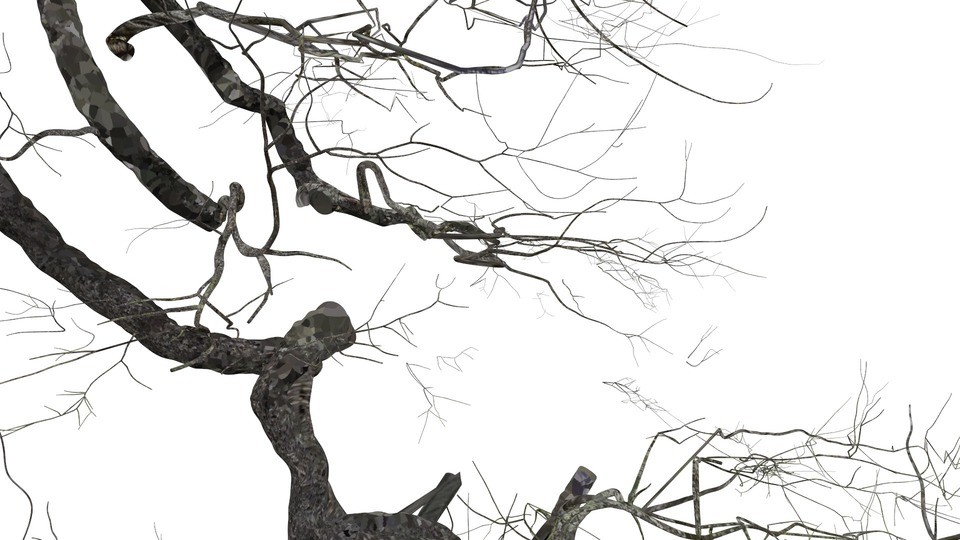}
    \twofigure{0.48\textwidth}{0.2in}{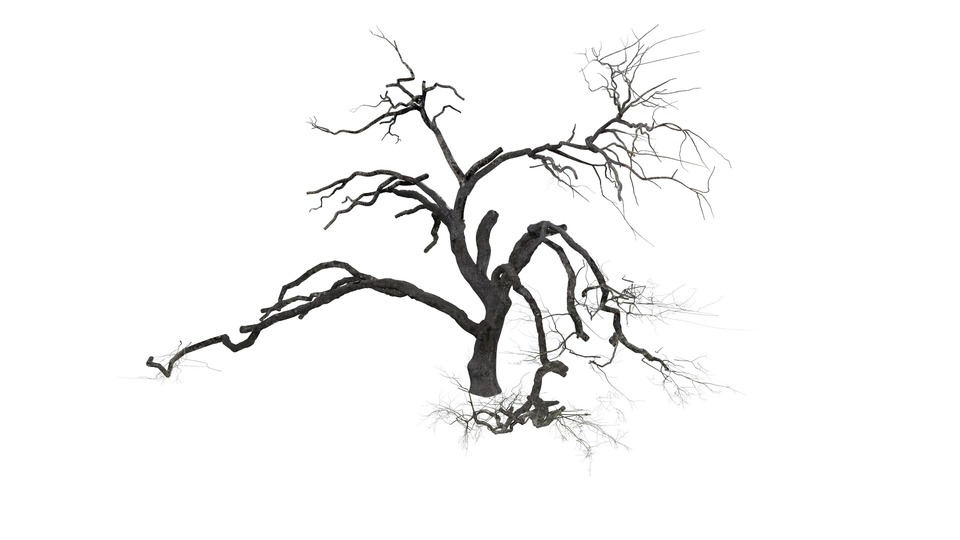}{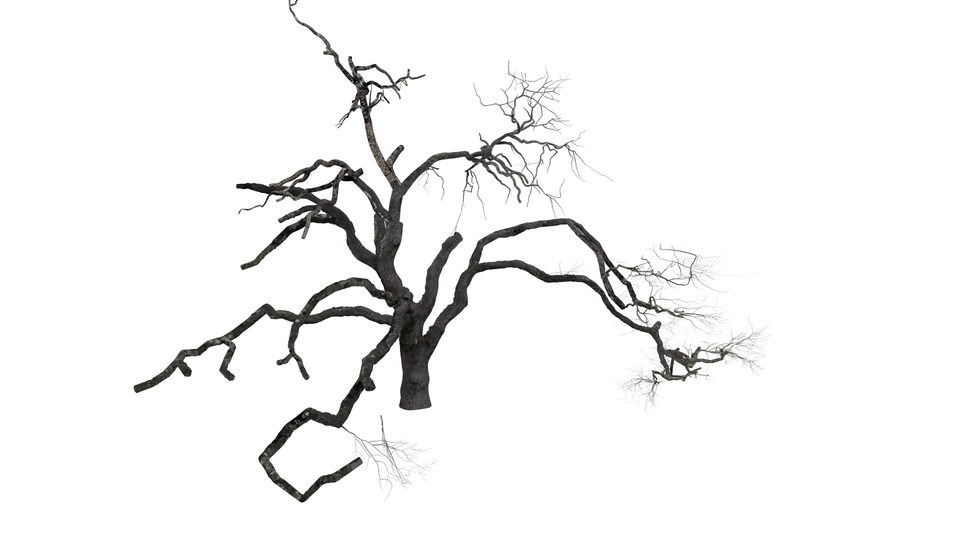}
    \caption{
        (Top row) Triangle mesh geometry perturbed using point cloud data.
        (Middle two rows) Medium scale branches generated from image annotations.
        (Bottom row) Zoomed out views of the final tree model (so far).
        \label{fig:mesh_views}}
\end{figure*}

Here we provide additional implementation details for the approach described in Section~\ref{sec:exp_med_branches}.
See Figure~\ref{fig:mesh_views} for additional views of the resulting meshed geometry.

\textbf{Misaligned Cameras: }
Some images may fail to be aligned properly after running structure from motion.
There are few enough of these that it is feasible to visually inspect each image and mark those with incorrect extrinsics as `misaligned.'
During medium scale branch reconstruction, we perform an initial pass of keypoint triangulation while withholding annotations from misaligned images.
We then use the resulting 3D triangulated keypoints to solve a perspective-\textit{n}-point problem for each misaligned image.
If this fails, then the annotations from that image are ignored; otherwise, the image is marked as `aligned.'
We then repeat the keypoint triangulation step using annotations from the newly aligned images along with those of the originally aligned images, thus potentially modifying the 3D positions of some keypoints while also potentially adding new keypoints.

\textbf{Keypoint Thickness Estimation: }

\begin{figure}[h]
    \centering
    \includegraphics[width=0.75\linewidth]{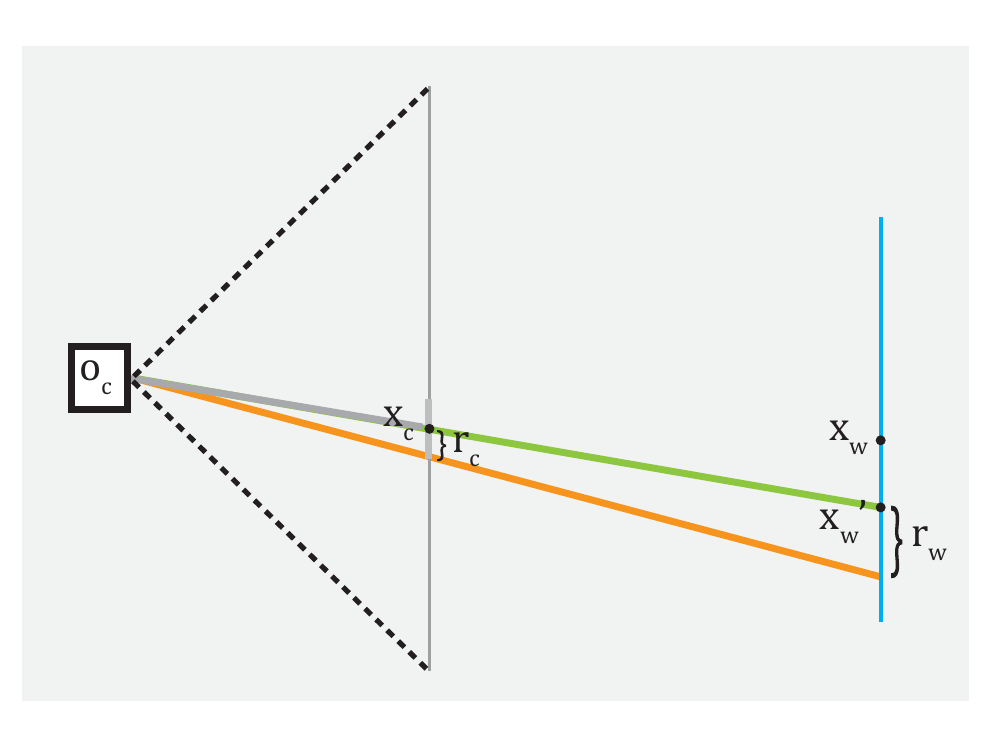}
    \vspace{-.2in}
    \caption*{}
    \label{fig:keypoint_radius}
    \vspace{-.2in}
\end{figure}

For a keypoint with triangulated position $x_w$, we estimate the world space radius by considering the set of cameras $C$ in which the keypoint is labeled.
For each such camera we know the camera's optical center $o_c$, the annotated point $x_c$, and the annotated radius $r_c$.
Let $p$ be the plane that contains $x_w$ and is parallel to the camera's image plane.
We intersect the ray $\overrightarrow{o_c x_c}$ with $p$ obtaining a point $x_w'$, then estimate the keypoint's radius $r_w$ using a ratio of distances: $\frac{r_w}{r_c} = \frac{||x_w'-o_c||}{||x_c-o_c||}$.
Finally, we average the radius estimates over each camera in which the keypoint is annotated so that the final radius estimate is $\frac{1}{|C|}\sum_{c\in C} \frac{||x_w'-o_c||}{||x_c-o_c||} r_c$.

\clearpage

\clearpage
{\small
\bibliographystyle{ieee}
\bibliography{references}
}

\end{document}